\documentclass[10pt]{article} 
\usepackage[accepted]{rlc}

\usepackage{amssymb}            
\usepackage{mathtools}          
\usepackage{mathrsfs}           
\usepackage{graphicx}           
\usepackage{subcaption}         
\usepackage[space]{grffile}     
\usepackage{url}                
\usepackage{wrapfig}
\usepackage[ruled,vlined]{algorithm2e}

\title{A Simple Mixture Policy Parameterization for Improving Sample Efficiency of CVaR Optimization}

\author{%
  Yudong Luo${^{1,4}}$, Yangchen Pan${^2}$, Han Wang${^3}$, Philip Torr${^2}$, Pascal Poupart${^{1,4}}$\\
  ${^1}$University of Waterloo, ${^2}$University of Oxford, ${^3}$University of Alberta, ${^4}$Vector Institute\\
}

\begin{document}

\maketitle

\begin{abstract}
Reinforcement learning algorithms utilizing policy gradients (PG) to optimize Conditional Value at Risk (CVaR) face significant challenges with sample inefficiency, hindering their practical applications. This inefficiency stems from two main facts: a focus on tail-end performance that overlooks many sampled trajectories, and the potential of gradient vanishing when the lower tail of the return distribution is overly flat. To address these challenges, we propose a simple mixture policy parameterization. This method integrates a risk-neutral policy with an adjustable policy to form a risk-averse policy. By employing this strategy, all collected trajectories can be utilized for policy updating, and the issue of vanishing gradients is counteracted by stimulating higher returns through the risk-neutral component, thus lifting the tail and preventing flatness. Our empirical study reveals that this mixture parameterization is uniquely effective across a variety of benchmark domains. Specifically, it excels in identifying risk-averse CVaR policies in some Mujoco environments where the traditional CVaR-PG fails to learn a reasonable policy. 
\end{abstract}

\vspace{-0.1in}
\section{Introduction}
\label{sec:intro}
\vspace{-0.05in}
Avoiding risks is a practical consideration in real world applications, inspiring risk-averse reinforcement learning (RL). Risk-averse RL involves optimizing some risk measures of the return random variable. Many risk measures have been studied, for instance, variance~\citep{tamar2012policy,la2013actor}, exponential utility functions~\citep{borkar2002q,fei2021risk}, Value at Risk (VaR)~\citep{chow2018risk,jung2022quantile}, and Conditional VaR (CVaR)~\citep{tamar2015optimizing,lim2022distributional}. We focus on CVaR in this work, which emphasizes the worst case outcome of a policy's return. Intuitively, CVaR measures the expected return below a specific quantile level $\alpha$, termed the risk level. This kind of risk is also known as the tail risk measure~\citep{liu2021theory}, since only the tail of a distribution is considered, and CVaR is more often preferred than VaR because it is coherent~\citep{delbaen2000coherent}.

Among the existing CVaR algorithms in RL~\citep{tamar2015optimizing,chow2018risk,tang2019worst,yang2021wcsac,ying2022towards}, policy gradient (PG) is a common choice. CVaR-PG samples a batch of $N$ trajectories and maximizes the mean return of the $\alpha N$ trajectories with worst returns~\citep{tamar2015optimizing}. This approach suffers from sample inefficiency due to two major facts~\citep{greenberg2022efficient}: 1) $1-\alpha$ portion of sampled trajectories are discarded; 2) gradients vanish when the tail of the return quantile function is overly flat, which is discussed later in Sec.~\ref{sec:cvarpg_problem}. Another line of research on optimizing CVaR is based on distributional RL~\citep{bellemare2017distributional}, e.g.,~\citet{dabney2018implicit,tang2019worst,keramati2020being}. However, due to the time-inconsistency of the risk, the objectives of some approaches differ from maximizing the $\alpha$-CVaR of the total return, while the behavior of some others are not well-understood yet~\citep{lim2022distributional}.

In this paper, we focus on the policy gradient approach and propose a simple mixture policy parameterization to improve sample efficiency. Our key insight is that in many real-world risk-sensitive domains, the agent may only need to perform risk-averse actions in a subset of states, e.g., related to risky regions, and behave akin to a risk-neutral agent in other states. We give an example in Sec.~\ref{sec:maze}. This motivates representing a risk-averse policy via integrating a risk-neutral policy and an adjustable component. With this parameterization, all collected trajectories can be used to update the policy under the mixture framework, and gradient vanishing is counteracted by stimulating higher returns with the help of its risk-neutral component,  thus lifting the tail and preventing flatness of the quantile function. To demonstrate the effectiveness of our method in learning risk-averse policies,  we modify several domains (Maze~\citep{greenberg2022efficient}, Lunar Lander\citep{brockman2016openai}, Mujoco~\citep{todorov2012mujoco}) where risk-aversion can be clearly verified. We empirically show that our method can learn a risk-averse policy when others fail to learn a reasonable policy. 

\textbf{Contributions.} To the best of our knowledge, a generally applicable approach to improve the sample efficiency of CVaR-PG algorithms remains unclear. In summary, our work provides 1) insights into a novel perspective on scenarios where risk-averse behavior is required only in a subset of states; 2) a simple mixture policy parameterization to address sample inefficiency. Notably, our algorithm, in certain Mujoco domains, marks a pioneering advancement in CVaR optimization.

\vspace{-0.05in}
\section{Background: CVaR Optimization in RL}
\label{sec:bg}
\vspace{-0.05in}
In standard RL settings, agent-environment interactions are modeled as a Markov Decision Process (MDP), represented as a tuple $\langle \mathcal{S},\mathcal{A}, P, R, \mu_0,\gamma \rangle$ \citep{puterman2014markov}. $\mathcal{S}$ and $\mathcal{A}$ denote state and action spaces. $P(\cdot|s,a)$ defines the transition. $R$ is the state and action dependent reward. $\mu_0$ is the initial state distribution, and $\gamma\in(0, 1]$ is a discount factor. An agent takes actions according to its policy $\pi:\mathcal{S}\times\mathcal{A}\rightarrow[0,+\infty)$. The return at time step $t$ is defined as $G^\pi_t=\sum_{i=0}^\infty \gamma^i R(s_{t+i},a_{t+i})$. Thus, $G^\pi_0$ is the random variable indicating the total return starting from the initial state following $\pi$. 
CVaR-based risk-averse RL avoids catastrophic outcomes by optimizing the tail risk measure of $G^\pi_0$, e.g., CVaR, instead of $\max_\pi \mathbb{E}[G^\pi_0]$ as done in risk-neutral RL.

\vspace{-0.03in}
\subsection{Problem Formulation}
\vspace{-0.03in}
Let $Z$ be a bounded random variable with cumulative distribution function $F_Z(z)=\mathcal{P}(Z\leq z)$. Denote the $\alpha$-quantile as $q_\alpha(Z) = \min\{z|F_Z(z)\geq\alpha\}, \alpha\in(0,1]$. The CVaR at confidence level $\alpha$ is given by~\citep{rockafellar2000optimization}
\vspace{-0.16in}
\begin{equation}
\label{eq:cvar-def-1}
    \mathrm{CVaR}_\alpha(Z)=\frac{1}{\alpha}\int_0^\alpha q_\beta(Z) d\beta
    \vspace{-0.05in}
\end{equation}
When $\alpha\rightarrow1$, $\mathrm{CVaR}_\alpha(Z)$ becomes $\mathbb{E}[Z]$. If $Z$ has a continuous distribution, $\mathrm{CVaR}_\alpha(Z)$ is more intuitively expressed as $\mathrm{CVaR}_\alpha(Z) = \mathbb{E}[Z|Z\leq q_\alpha(Z)]$. Thus, $\mathrm{CVaR}_\alpha(Z)$ can be interpreted as the expected value of the $\alpha$-portion of the left tail of the distribution of $Z$. Another way to define $\mathrm{CVaR}_\alpha(Z)$ is \citep{rockafellar2000optimization}
\vspace{-0.1in}
\begin{equation}
\label{eq:cvar_def-2}
    \mathrm{CVaR}_\alpha(Z) = \max_{k\in\mathbb{R}} k - \frac{1}{\alpha}\mathbb{E}[(k-Z)^+]
    \vspace{-0.05in}
\end{equation}
\vspace{-0.03in}
where $(x)^+=\max\{x,0\}$, and the maximum is always attained at $k=q_\alpha(Z)$ as a by product.

In this paper, we consider the problem of maximizing the CVaR of total return $G^\pi_0$ given a confidence level $\alpha$ (we consider small $\alpha$ in practice)~\citep{tamar2015optimizing}, i.e.,
\vspace{-0.04in}
\begin{equation}
    \label{eq:objective}
    \max_\pi \mathrm{CVaR}_\alpha(G^\pi_0)
    \vspace{-0.05in}
\end{equation}
\textbf{Remark}. Some works optimize the CVaR term plus the mean term or treat CVaR as a constraint~\citep{chow2018risk,yang2021wcsac,ying2022towards}, which differ from the problem in Eq.~\ref{eq:objective}. In addition, the risk defined on the total return (Eq.~\ref{eq:objective}) is known as the static risk. Another line of research on CVaR works on dynamic risk \citep{ruszczynski2010risk,huang2021convergence,du2023provably}, where risk is recursively computed at each time step. The comparison between static and dynamic CVaR is discussed, e.g., in~\citet{lim2022distributional}. 
\vspace{-0.06in}
\subsection{CVaR Policy Gradient (CVaR-PG)}
\vspace{-0.03in}
\label{sec:cvar-pg}
Consider $\pi$ is parameterized by $\theta$. Under some mild assumptions, the gradient of  Eq.~\ref{eq:objective} w.r.t. $\theta$ can be estimated by sampling trajectories $\{\tau_i\}_{i=1}^N$ from the environment using $\pi_\theta$ \citep{tamar2015optimizing}.
\vspace{-0.15in}
\begin{equation}
\label{eq:cvar-pg}
\nabla_\theta \mathrm{CVaR}_\alpha (G^{\pi_\theta}_0)\simeq \frac{1}{\alpha N} \sum_{i=1}^N \mathbb{I}_{\{R(\tau_i)\leq \hat{q}_\alpha\}} (R(\tau_i)-\hat{q}_\alpha) \sum_{t=0}^T \nabla_\theta \log\pi_\theta(a_{i,t}|s_{i,t})
\vspace{-0.05in}
\end{equation}
where $R(\tau)$ represents the total return of trajectory $\tau$, $\hat{q}_\alpha$ is the empirical $\alpha$-quantile estimated from $\{R(\tau_i)\}_{i=1}^N$, and $T$ is the maximum trajectory length. This gradient is derived from Eq.~\ref{eq:cvar-def-1}. For further details, readers can refer to \citet{tamar2015optimizing}. Note that computing policy gradient from Eq.~\ref{eq:cvar_def-2} is also feasible and results in a similar update as Eq.~\ref{eq:cvar-pg}, e.g., see Algo.~1 in \citet{chow2018risk}.
\vspace{-0.03in}
\subsection{Distributional RL with CVaR}
\label{sec:drl}
\vspace{-0.03in}
Distributional RL \citep{bellemare2017distributional} is recently used for CVaR optimization. Since it directly learns a value distribution, the risk metric is easy to compute. Denote the return random variable at the state-action pair $(s,a)$ as $Z^\pi(s,a)=\sum_{t=0}^\infty\gamma^t R(s_t,a_t)$, where $s_0=s$, $a_0=a$, $s_{t+1}\sim P(\cdot|s_t,a_t)$, and $a_t\sim\pi(\cdot|s_t)$. Then the distributional Bellman equation is given by $Z^\pi(s,a)\overset{D}{=} R+\gamma Z^\pi(S',A')$, with $S'\sim P(\cdot|s,a)$, $A'\sim \pi(\cdot|S')$, and $X\overset{D}{=}Y$ indicates that random variables $X$ and $Y$ follow the same distribution. The well known $Q$-value can be extracted by $Q^\pi(s,a)=\mathbb{E}[Z^\pi(s,a)]$. 

\citet{dabney2018implicit,keramati2020being} propose to select actions according to
\vspace{-0.07in}
\begin{equation}
\label{eq:mkv}
    Z^\pi(s,a)\overset{D}{=} R+\gamma Z^\pi(S',A'),~~~~A' = \arg\max_{a'}\mathrm{CVaR}_\alpha(Z^\pi(S',a'))
    \vspace{-0.06in}
\end{equation}
This strategy always selects actions leading to the largest $\alpha$-CVaR at the current step and is termed as "Markov action selection strategy" by \citet{lim2022distributional}. Within the framework of actor critic, a similar way is applied by updating the actor towards the $\alpha$-CVaR of the critic, e.g., see \cite{tang2019worst}. However, \citet{lim2022distributional} showed this strategy converges to neither static nor dynamic optimal CVaR policies by counterexamples. Thus, it is not consistent with the problem in Eq.~\ref{eq:objective}.

\citet{bauerle2011markov} simplified Eq.~\ref{eq:cvar_def-2} by avoiding optimizing $k$ and fixing it to some constant $k_0$, resulting in the problem $\max_\pi -\mathbb{E}[(k_0-G^\pi_0)^+]$. This problem can be modeled by an augmented MDP with new state $\Tilde{s}=(s,k)\in\mathcal{S}\times\mathbb{R}$, where $k$ is a moving variable keeping track of the accumulated rewards so far. \citet{lim2022distributional} incorporated this perspective with distributional RL by introducing the tracking variable, and proposed a new action selection strategy as 
\vspace{-0.07in}
\begin{equation}
\label{eq:lim}
    Z^\pi(s,a)\overset{D}{=} R+\gamma Z^\pi(S',A'),~~~~A' = \arg\max_{a'} \mathbb{E}[-(\frac{k-R}{\gamma}-Z^\pi(S',a'))^+]
    \vspace{-0.06in}
\end{equation}
where $k$ is the tracking variable at $(s,a)$, and is set to $\alpha$-CVaR for the initial state. \citet{lim2022distributional} showed that the optimal CVaR policy is a fixed point of Eq.~\ref{eq:lim} if it exists and it is unique. However, when $\pi$ is not CVaR optimal, its behavior is generally unknown.
\vspace{-0.03in}
\subsection{Other CVaR RL Algorithms}
\vspace{-0.03in}
There are several other CVaR algorithms in the context of MDPs, where full knowledge of the MDP is required. Thus, they are not applicable to RL problems where transition dynamics are unknown. These works are less relevant to ours and hence we only provide a brief review here. Based on the theory of CVaR decomposition \citep{pflug2016time}, a dynamic programming approach is developed by decomposing the CVaR via its risk envelope \citep{chow2015risk}. This approach returns the optimal $\alpha$-CVaR value for any $\alpha\in(0,1]$. Recently, \citet{hau2023dynamic} pointed out this method has some flaws in the control setting, and provided counter examples.

\vspace{-0.03in}
\section{Mixture Parameterization Policies}
\vspace{-0.05in}
In this section, we examine the difficulties inherent in classical CVaR-PG methods. This examination sets the stage for our proposed solution: a mixture parameterization approach.
\vspace{-0.04in}
\subsection{Challenges of CVar-PG: low-efficiency gradient estimation}
\label{sec:cvarpg_problem}
\vspace{-0.03in}
The classical CVaR-PG (Eq.~\ref{eq:cvar-pg}) faces two significant challenges that undermine its sample efficiency and practical applicability. Firstly, to emphasize the tail outcomes, a small value of $\alpha$ is chosen. Consequently, only an $\alpha$-fraction of the trajectories contribute to the gradient estimation in Eq.~\ref{eq:cvar-pg}, leading to the discarding of the majority of trajectories and resulting in low sample efficiency. 

\begin{figure}[t]
\vspace{-0.1in}
    \begin{center}
        \includegraphics[width=0.85\textwidth]{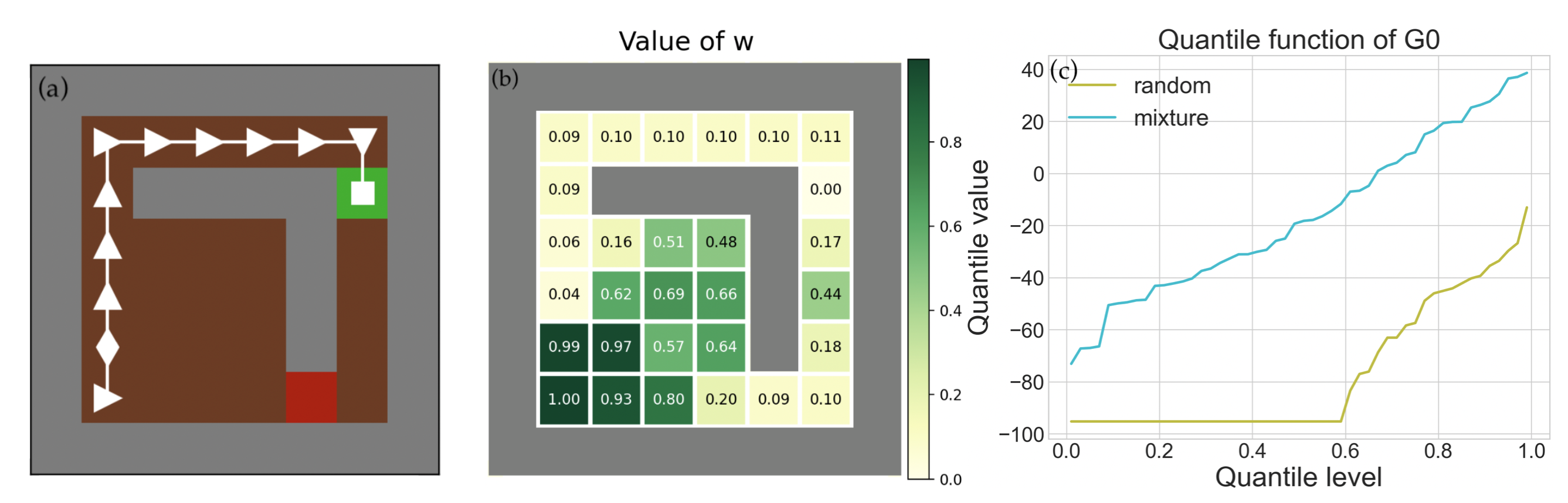}
    \end{center}
    \caption{(a) A maze domain with green goal state. The red state returns an uncertain reward (details in Sec.~\ref{sec:maze}). Triangle pointers indicate the risk-neutral actions (not unique for the second state). (b) Value of $w$ of Eq.~\ref{eq:mix-pi} for each state after the mixture policy is updated by CVaR-PG. (c) The empirical quantile function of the total return in maze at an early training stage, if the initial policy is a random and mixture policy.}
    \label{fig:maze-all}
    \vspace{-0.15in}
\end{figure}

Secondly, as identified by \citet{greenberg2022efficient}, a small $\alpha$ also introduces a gradient vanishing issue. This occurs because the term $\mathbb{I}_{\{R(\tau_i)\leq \hat{q}_\alpha\}} (R(\tau_i)-\hat{q}_\alpha)$ can equal zero for any $\tau_i$ satisfying $R(\tau_i)\leq \hat{q}_\alpha$, i.e., $R(\tau_i)=\hat{q}_\alpha$ for those trajectories $\tau_i$ selected by the indicator function. This issue arises when the left tail of the quantile function is notably flat, meaning that all quantile values below the $\alpha$-quantile are identical. Such a scenario is particularly likely in environments with a discrete rewards distribution, a fact that is often overlooked when assuming continuous rewards. For illustration, we present the empirical quantile function of $G^\pi_0$ obtained through Monte Carlo sampling in Fig.~\ref{fig:maze-all}(c), during the initial training phase with a random policy in a maze environment (detailed in Sec.~\ref{sec:maze} and shown in Fig.~\ref{fig:maze-all}(a)). In this scenario, if the agent neither reaches the goal nor enters the red state, the resulting trajectories will yield identical low returns, leading to a markedly flat left tail of the quantile function for $G^\pi_0$.

To tackle gradient vanishing, \citet{greenberg2022efficient} proposed curriculum learning by starting from an $\alpha$ close to $1$ (risk-neutral) and gradually decreasing $\alpha$ to its target value. To further improve sample efficiency, \citet{greenberg2022efficient} proposed a sampling method based on cross-entropy to sample high-risk scenarios from the environment. The algorithm is then focused on learning high-risk parts of the environment and thus improving sample efficiency. However, this sampling strategy requires knowledge of the environment dynamics and the ability to control the parameters of the dynamics in ways that are domain specific, which is not realistic for many RL domains.
\vspace{-0.04in}
\subsection{Mixture with Risk-neutral Policy}
\label{sce:mix-with-neutral}
\vspace{-0.03in}

To address the aforementioned challenges, our key observation is that many real-world risk-sensitive applications exhibit a pattern wherein only a subset of states requires risk-averse behavior. In the remaining portion of the state space, the agent can behave akin to a risk-neutral agent. For example, in scenarios with minimal or no other cars on a highway, a driver may simply need to follow the road without slowing down or braking, as long as the vehicle remains under the speed limit. This observation leads us to propose representing the policy as a mixture of a risk-neutral policy and an adjustable component, i.e.,
\vspace{-0.05in}
\begin{equation}
\label{eq:mix-pi}
    \pi(a|s) = w(s) \pi'(a|s) + (1-w(s))\pi^n(a|s)
    \vspace{-0.05in}
\end{equation}
where $w(s)\in[0,1]$ is the mixture component weight. $\pi^n$ is the risk neutral policy, and $\pi'$ is the adjustable policy. At different phases of a task, the agent self-selects the most suitable policies to execute to ensure the overall policy $\pi$ is risk averse. 

It is evident that the proposed parameterization effectively addresses the challenges outlined earlier. Firstly, it allows for the use of all trajectories collected so far to update the risk-neutral policy within the mixture framework. Secondly, the risk-neutral component encourages the agent to venture into areas of high reward, potentially avoiding the flat tail of the return distribution, and hence mitigates the issue of vanishing gradients. We illustrate the advantages in the following example.

\vspace{-0.03in}
\subsection{A Motivating Maze Example}
\label{sec:maze}
\vspace{-0.04in}
Consider a maze domain in Fig.~\ref{fig:maze-all}(a), which is originally from \citet{greenberg2022efficient} and slightly modified by \citet{luo2024alternative}. Starting from the bottom left corner, the goal of the agent is to reach the green goal state. The gray color marks the walls. The per-step reward is deterministic (i.e., -1) except for the red state, whose reward distribution is $-1+\mathcal{N}(0,1)\times 30$. The reward for visiting the goal is a positive constant value (i.e., 10). Thus, the shortest path going through the red state towards the goal is the optimal risk-neutral path, while the longer path (shown in white color) is $\alpha$-CVaR optimal if $\alpha$ is small, though its expected return is slightly lower.

In this domain, suppose we are given the optimal risk neutral policy for each state (which is actually easy to get, e.g., via Q-learning or value iteration \citep{sutton1998introduction}, or even by observing the shortest path), it is easy to see most actions along the white (i.e., risk-averse) path are the same as the risk-neutral policy except the initial state. This means the risk-averse agent only needs to adjust the actions at that state and then follow the optimal risk-neutral policy afterwards. We validate this idea by visualizing the value of $w(s)$ of Eq.~\ref{eq:mix-pi} in Fig.~\ref{fig:maze-all}(b), after the mixture policy is trained by CVaR-PG. The value of $w(s)$ represents the probability of choosing $\pi'$ at each state. Here the risk neutral policy $\pi^n$ is pre-computed and provided as the softmax of the optimal $Q$-values (we use temperatures to make the entropy of $\pi^n$ small). Thus, $\pi'$ and $w$ are the components that need to be learned by CVaR-PG. As shown in the figure, the probability of choosing $\pi'$ is only high in the surroundings of the starting state, and the probability of choosing $\pi^n$ significantly increases after the agent moves far away from the beginning. Also, the empirical quantile function of $G^\pi_0$ obtained by this mixture policy at the initial training phase is shown in Fig.~\ref{fig:maze-all}(c). Compared with the randomly initialized policy, the flat tail is eliminated, thereby preventing gradient vanishing.

\begin{wrapfigure}{r}{0.5\textwidth}
\vspace{-0.2in}
    \begin{algorithm}[H]
    \label{algo:mix-}
        \caption{Mixture policy for CVaR-PG}
        \KwIn{risk level $\alpha$, trajectories sampled per batch $N$, training steps $M$, IQL update frequency $C$}
        \textbf{Initialize:}  policy $\pi_\theta=w_{\theta_2}\pi'_{\theta_1}+(1-w_{\theta_2})\pi^n_\vartheta$ where $\theta=(\theta_1,\theta_2)$, buffer $B$, Q-function $Q_\phi$ (target $Q_{\hat{\phi}}$), value function $V_\psi$\;
        \For{$m$ in $1:M$}{
            \tcp{\textcolor{blue}{Sample trajectories}}
            $\{\tau_i\}_{i=1}^N \leftarrow$ run\_episodes($\pi_\theta,N$) \;
            Store $\{\tau_i\}_{i=1}^N$ to $B$\;
            \tcp{\textcolor{blue}{CVaR PG, i.e., Eq.~\ref{eq:cvar-pg}} }
            Update $\theta$ via CVaR-PG($\pi_\theta, \{\tau_i\}_{i=1}^N,\alpha$)\;
            \tcp{\textcolor{blue}{Risk-neutral, e.g., IQL updates}}
            \If{m \% C == 0}{
            Sample $\mathcal{D}=\{(s,a,r,s')\}\sim B$\;
            Update $Q_\phi$ via Eq.~\ref{eq:iql-q}\;
            Update $V_\psi$ via Eq.~\ref{eq:iql-v}\;
            Update $\pi^n_\vartheta$ via Eq.~\ref{eq:iql-pi}\;
            }
        }
    \end{algorithm}
\vspace{-0.6in}
\end{wrapfigure}
\textbf{Remark.} This concept, where risk-averse behavior is required only in a subset of states, extends to various fields. For example, in portfolio management, such behavior is crucial only in particular market trends~\citep{ji2019risk,yu2023learning}, and in healthcare, it is essential only with specific health indicator warnings~\citep{mulligan2023risk}.

\vspace{-0.03in}
\subsection{Offline RL Risk Neutral Learning}
\label{sec:learning}
\vspace{-0.03in}
This section explores the process of acquiring a risk-neutral policy under the function approximation setting. We discovered that incorporating a pre-trained deep risk-neutral policy into the mixture policy frequently leads to a suboptimal risk-averse policy, a finding that is further elaborated in Appendix~\ref{sec:pre-train-risk-neutral}.

Observing that the update of CVaR-PG typically involves collecting substantial trajectories, these trajectories naturally constitute an empirical MDP to which an offline RL algorithm can be applied to extract a risk-neutral policy. The field of offline RL has seen rapid advancements in recent years, offering promising solutions for solving the empirical MDP formed from the collected trajectories.

Offline RL attempts to learn an optimal policy from a pre-gathered offline dataset $\mathcal{D}=\{(s,a,s',r)\}_{i=1}^n$, where the learning algorithm is restricted to learning from the samples contained within $\mathcal{D}$ without any additional interaction with the real environment. One key challenge in offline RL is to not overestimate the action values outside of the dataset~\citep{fujimoto2019offline}. To address this challenge, there are generally two strategies. The first approach aims to keep the learned policy closely aligned with the dataset's policy by applying some KL constraint, ensuring the learned policy remains within the dataset's support~\citep{peng2020advantage,brandfonbrener2021offline,fujimoto2021minimalist}. The second strategy involves directly optimizing the policy using the samples available in the dataset~\citep{fujimoto2019offline,kostrikov2022offline,xiao2023offline}.

In our research, we utilize Implicit $Q$-Learning (IQL)~\citep{kostrikov2022offline} for learning risk-neutral policies, chosen for its proven reliability and empirical validation. 
IQL possesses a $Q$ estimator $Q_\phi(s,a)$, a value estimator $V_\psi(s)$, and a policy $\pi^n_{\vartheta}(a|s)$. $Q$-function is updated via minimizing
\vspace{-0.04in}
\begin{equation}
\label{eq:iql-q}
    L_Q(\phi)=\mathbb{E}_{(s,a,s')\sim\mathcal{D}}[(r(s,a)+\gamma V_\psi(s')-Q_\phi(s,a))^2]
    \vspace{-0.04in}
\end{equation}

Value function is updated via expectile regression to avoid overestimation ($Q_{\hat{\phi}}$ is the target function)
\vspace{-0.03in}
\begin{equation}
\label{eq:iql-v}
    L_V(\psi)=\mathbb{E}_{(s,a)\sim\mathcal{D}}[L_2^\eta(Q_{\hat{\phi}}(s,a)-V_\psi(s))],~~ L_2^\eta(u)=|\eta-\mathbb{I}_{\{u<0\}}|u^2
    \vspace{-0.03in}
\end{equation}

Policy is updated by advantage-weighted regression~\citep{peters2007reinforcement} with temperature $\beta$
\vspace{-0.04in}
\begin{equation}
\label{eq:iql-pi}
    L_{\pi^n}(\vartheta)=\mathbb{E}_{(s,a)\sim\mathcal{D}}[\exp(\beta(Q_{\phi}(s,a)-V_\psi(s)))\log\pi^n_{\vartheta}(a|s)]
    \vspace{-0.04in}
\end{equation}

All the trajectories are stored in a replay buffer for IQL update to learn $\pi^n$. In practice, we can perform this update after enough transition data are collected. The overall process of training the mixture policy is described in Algo.~\ref{algo:mix-}.

\vspace{-0.05in}
\section{Experiments}
\label{sec:exp}
\vspace{-0.05in}
We modify several domains such that the risk-averse behavior is clear to identify to evaluate the algorithms. We include REINFORCE with baseline method, as a risk-neutral baseline. In more complex domains, we use SAC~\citep{haarnoja2018soft} instead. 

\begin{wrapfigure}{r}{0.5\textwidth}
\vspace{-0.3in}
  \begin{center}
    \includegraphics[width=0.5\textwidth]{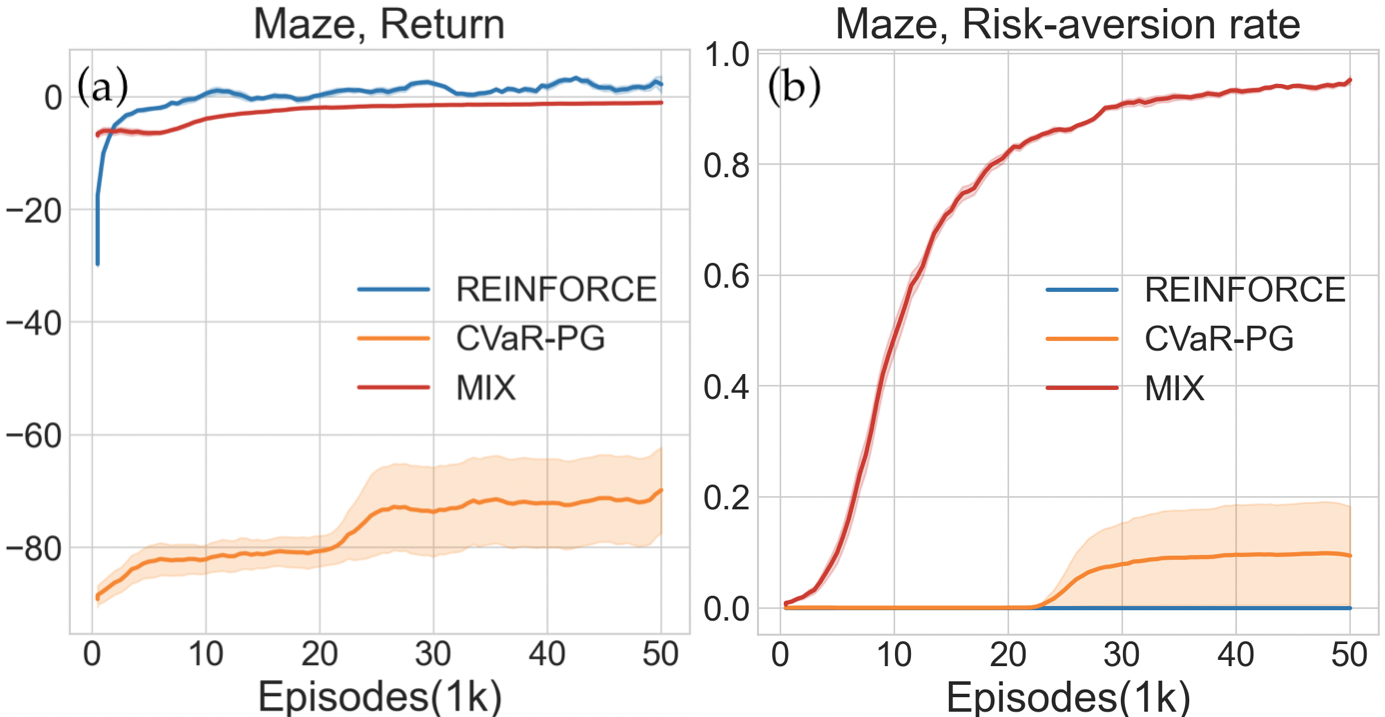}
  \end{center}
  \vspace{-0.1in}
  \caption{(a) Policy return (y-axsis) and (b) Risk-aversion (long path) rate (y-axsis) v.s. training episodes in Maze. Curves are averaged over 10 seeds with shaded regions indicating standard errors.}
  \label{fig:maze-res}
  \vspace{-0.15in}
\end{wrapfigure}

\textbf{Baselines.} We compare our method with CVaR-PG in Eq.~\ref{eq:cvar-pg}~\citep{tamar2015optimizing}, distributional RL with Markov action selection strategy in Eq.~\ref{eq:mkv} (denoted as DRL-mkv), and Lim's action selection strategy in Eq.~\ref{eq:lim}~\citep{lim2022distributional} (denoted as DRL-lim). In continuous action domains, we adapt DRL-mkv and DRL-lim by DPG~\citep{silver2014deterministic} as done in~\citet{tang2019worst}, see Sec.~\ref{sec:baseline-info} for an overview. We use MIX to represent our method. Pre-computed $\pi^n$ in maze is provided to MIX as described in Sec~\ref{sec:maze} since it is easy to get. In other domains, $\pi^n$ is learned by IQL during training. Please refer to Appendix~\ref{sec:experiments} for any missing implementation details.

\textbf{Remark.} The method in~\citet{greenberg2022efficient} is CVaR-PG with curriculum learning and a special trajectory sampling strategy, which is orthogonal to our approach. It requires to control the environment dynamics, and may not be straightforwardly applicable to most domains discussed here.
We compare with it in one domain from ~\citet{greenberg2022efficient} in Sec.~\ref{sec:driving}.
\vspace{-0.05in}
\subsection{Tabular case: Maze Problem}
\vspace{-0.03in}

This domain is modified from~\citet{greenberg2022efficient} that was previously described in Sec.~\ref{sec:maze}. The maximum episode length is $100$. CVaR $\alpha=0.1$.  REINFORCE, CVaR-PG, and MIX collect $N=50$ episodes before updating the policy. Here we report the rate of choosing the long path during training in Fig.~\ref{fig:maze-res}(b). Since the policy is  non-deterministic, the length of the sampled risk-averse path may not be exactly $11$ (the length of the white path in Fig.~\ref{fig:maze-all}(a)). Here we treat a path as risk-averse if it goes towards the top, reaches the goal, and the path length does not exceed $14$.

CVaR-PG fails to learn a reasonable policy even in this simple domain due to gradient vanishing as discussed in Sec.~\ref{sec:cvarpg_problem}. We show the gradient norm of CVaR-PG in Fig.~\ref{fig:cvarpg-norm} in appendix to further illustrate this phenomenon in Maze. By initializing MIX with a risk neutral policy, it achieves a relatively high return at the early learning phase, thus potentially avoids gradient vanishing.
\vspace{-0.04in}
\subsection{Discrete control: LunarLander}
\vspace{-0.03in}
\begin{figure}[ht]
\vspace{-0.05in}
    \begin{center}
        \includegraphics[width=0.97\textwidth]{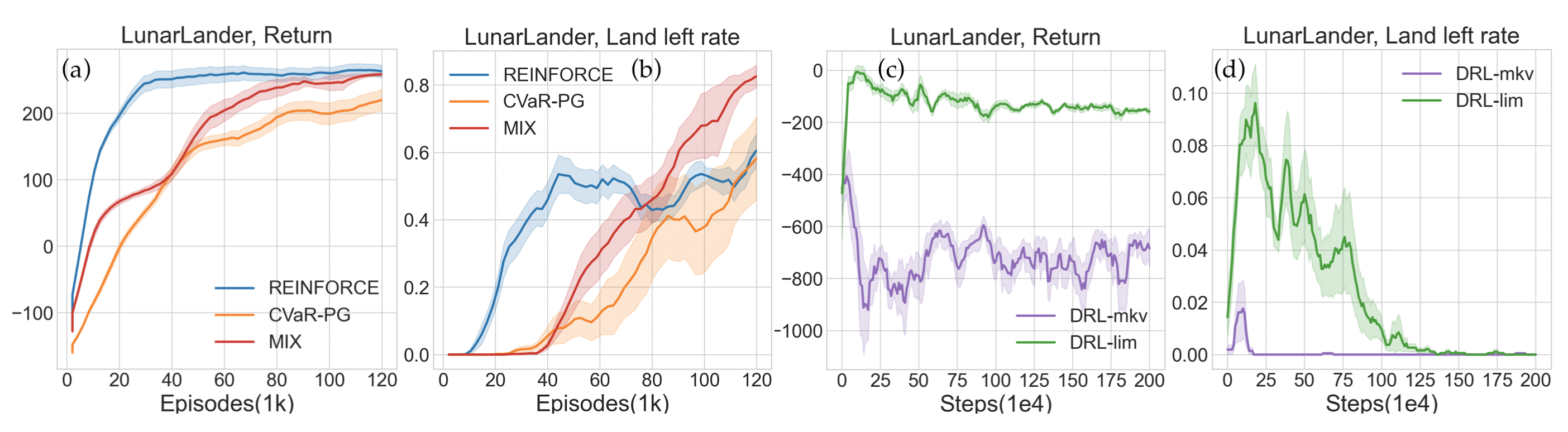}
    \end{center}
    \caption{(a,c) Policy return (y-axis), and (b,d) Left-landing rate (i.e., risk-averse landing rate) (y-axis) v.s. training episodes or steps in LunarLander. Curves are averaged over 10 seeds with shaded regions indicating standard errors. For the landing left rate, higher is better.}
    \label{fig:ll-res}
    \vspace{-0.06in}
\end{figure}

This domain is taken from OpenAI Gym~\citep{brockman2016openai}. We refer readers to its official documents for a full description. The goal of the agent is to land on the ground without crashing. We split the ground into left and right parts by the middle line of the landing pad, as shown in Fig.~\ref{fig:LL} in appendix. If landing on the right, an additional noisy reward sampled from $\mathcal{N}(0,1)$ times 100 is given. A risk-averse agent should learn to land on the left as much as possible. We set CVaR $\alpha=0.1$. REINFORCE, CVaR-PG, and MIX collect $N=30$ episodes before updating the policy. DRL-mkv and DRL-lim are off-policy methods and update policies at each environment step. We train them for 2e6 steps instead of as many episodes as other methods.

We report the left-landing rate of different methods in Fig.~\ref{fig:ll-res}(b) and (d). Comparing DRL-mkv and DRL-lim against episode-based algorithms is not straightforward within the same figure due to the difference in parameter update frequency. Thus we show them separately. MIX achieves a comparable return with REINFORCE at the end, but shows a clear risk-aversion by landing more on the left. DRL-mkv and DRL-lim can not learn a reasonable policy given the small CVaR $\alpha$.  As mentioned in Section~\ref{sec:drl}, they optimize a different objective than CVaR that is not well understood.

\begin{figure}[t]
\vspace{-0.04in}
    \begin{center}
        \includegraphics[width=0.97\textwidth]{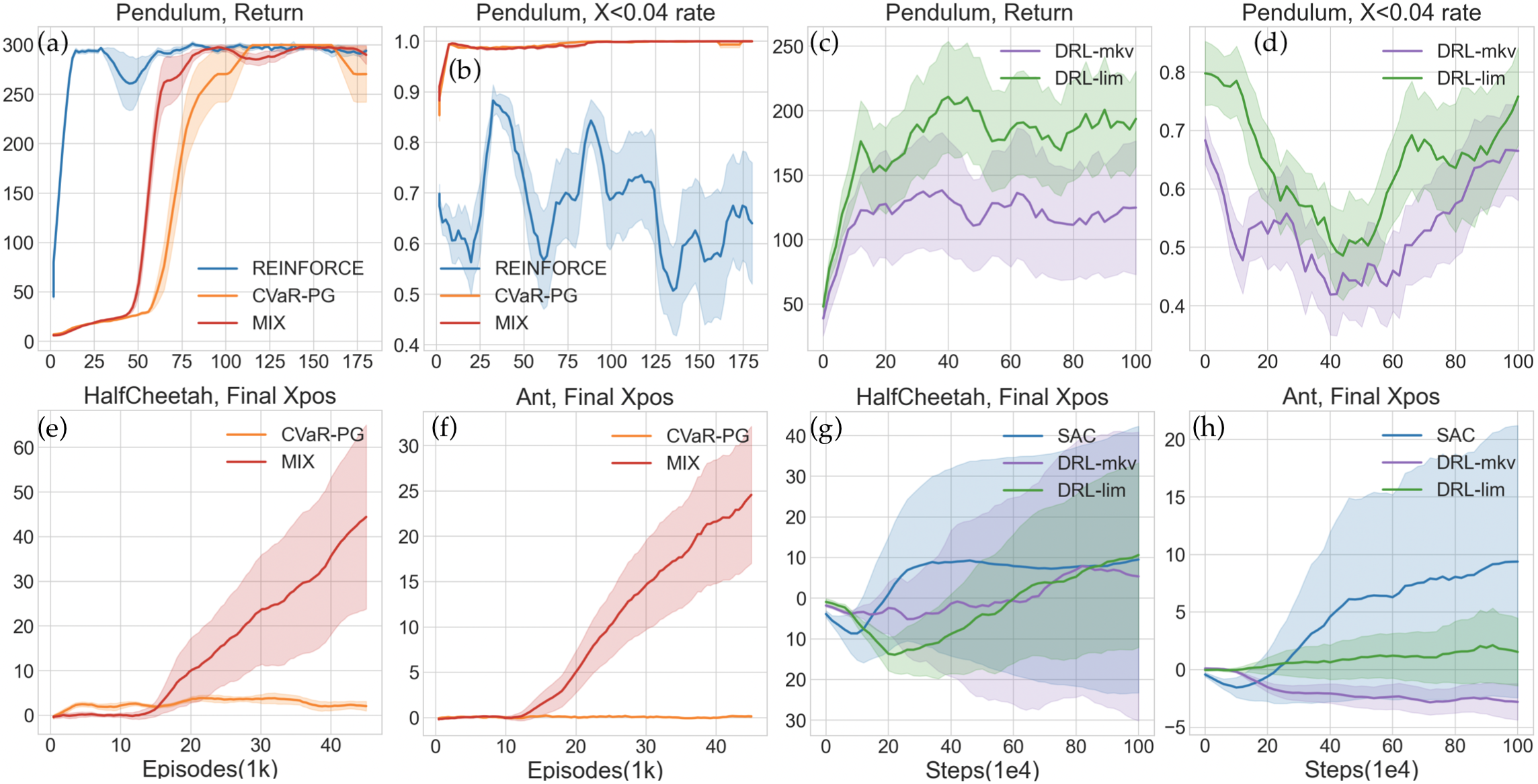}
    \end{center}
    \caption{(a, c) Policy return (y-axsis) in InvertedPendulum, (b, d) visiting non-noisy region rate (y-axis) in InvertedPendulum, (e, g) Final X-position (y-axsis) in HalfCheetah, (f, h) Final X-position in Ant (y-axsis) v.s. training episodes or steps in Mujoco. Curves are averaged over 10 seeds with shaded regions indicating standard errors. For the location visiting rate, higher is better.}
    \label{fig:mujoco-res}
    \vspace{-0.09in}
\end{figure}
\vspace{-0.04in}
\subsection{Continuous control: Mujoco}
\vspace{-0.04in}
Mujoco~\citep{todorov2012mujoco} is a collection of robotics environments with continuous states and actions in OpenAI Gym~\citep{brockman2016openai}. Here, we select three domains, namely InvertedPendulum, HalfCheetah, and Ant (disparities in the inherent difficulty of moving backward versus forward are observed in other domains, which likely stems from the intrinsic design of the physics engine’s dynamics). Inspired by \citet{malik2021inverse,liu2022benchmarking}, we define the risky region based on the X-position. Specifically, if X-position $>0.04$ in InvertedPendulum, X-position $<-3$ in HalfCheetah and Ant, a zero-mean Gaussian noise is added to the reward ($\mathcal{N}(0,1)\times 10$ in InvertedPendulum, $\mathcal{N}(0,1)\times 50$ in HalfCheetah and Ant). To further ensure that agents move both forward and backward with equal preference in terms of expected reward in the two environments, we define the distance-based reward as the difference in distance between the current and previous states from the origin, regardless of the sign of the X-position. 
Consequently, in InvertedPendulum, a risk-averse agent aims to keep the pendulum balanced while staying out of the noisy region. In HalfCheetah and Ant, a risk-averse agent should learn to move toward the opposite direction of the noisy region. We optimize CVaR $\alpha=0.2$. REINFORCE still serves as the risk neutral baseline in InvertedPendulum. In HalfCheetah and Ant, we use SAC~\citep{haarnoja2018soft} instead, since the vanilla policy gradient is not good at more complex domains. REINFORCE, CVaR-PG, and MIX collect $N=30$ episodes before updating the policy in InvertedPendulum, and $N=15$ in HalfCheetah and Ant. DRL-mkv, DRL-lim, and SAC are trained for 1e6 steps.

We report the total return and X<0.04 rate in InvertedPendulum, which are sufficient to reflect the risk-averse behavior of the agent, since the reward is 1 as long as the pendulum is balanced. In HalfCheetah and Ant, we report the final X-position in Fig.~\ref{fig:mujoco-res}, as the return can not reflect which direction the agent is moving in the two domains. The policy returns in these two domains are shown in Fig.~\ref{fig:hc-ret} and \ref{fig:ant-ret} in appendix. CVaR-PG achieves a risk-averse policy in InvertedPendulum, i.e., high return and high rate of staying in the non-noisy region. But it fails to learn a reasonable policy in HalfCheetah and Ant, i.e., the final X-position is always close to the origin. MIX learns risk-averse policy by moving away from the noisy region. DRL-mkv and DRL-lim generally do not work well in all three domains since they optimize a different objective than CVaR that is not well understood (see Sec~\ref{sec:drl}).

\vspace{-0.08in}
\section{Conclusions and Future Work}
\vspace{-0.08in}
This paper proposes a mixture policy framework for CVaR-PG. It is motivated to overcome the sample inefficiency of the original CVaR-PG, caused by the waste of most sampled trajectories and gradient vanishing in some domains. We empirically show that our method can succeed when others fail to learn a risk-averse or a reasonable policy by mitigating the sample efficiency issue.

\textbf{Limitations and future work.} We have pinpointed a class of risk-averse RL problems characterized by requiring risk-averse behavior in a subset of states, suitable for our mixture approach. Though this category intuitively covers a broad range of scenarios, situations that do not fit this framework remain unexplored in this paper. Additionally, our method can be  potentially integrated with other techniques aimed at enhancing sample efficiency, e.g.,~\citet{greenberg2022efficient}, given its versatile nature. However, exploring such hybrid methodologies falls outside the scope of our current research. Observing the two limitations, researching algorithms to enhance the sample efficiency for the broader class of risk-averse problems, as well as possible integration with existing methods to improve sample efficiency, remains valuable for future work.

\subsubsection*{Broader Impact Statement}
\label{sec:broaderImpact}
This paper presents work whose goal is to advance the risk-averse reinforcement learning. There may be potential societal consequences of our work, none of which we feel must be specifically highlighted here.

\subsubsection*{Acknowledgments}
\label{sec:ack}

Resources used in this work were provided, in part, by the Province of Ontario, the Government of Canada through CIFAR, companies sponsoring the Vector Institute
(https://vectorinstitute.ai/partners/), the Natural Sciences and Engineering Council
of Canada, and the Digital Research Alliance of Canada (alliancecan.ca). Yudong Luo is also supported by a David R. Cheriton Graduate Scholarship, a President’s
Graduate Scholarship, and an Ontario Graduate Scholarship. Yangchen Pan would like to acknowledge support from Turing World leading fellow. Philip Torr is supported by the UKRI grant: Turing AI Fellowship. Philip Torr would also like to thank the Royal Academy of Engineering and FiveAI.

\bibliography{main}

\begin{thebibliography}{58}
\providecommand{\natexlab}[1]{#1}
\providecommand{\url}[1]{\texttt{#1}}
\expandafter\ifx\csname urlstyle\endcsname\relax
  \providecommand{\doi}[1]{doi: #1}\else
  \providecommand{\doi}{doi: \begingroup \urlstyle{rm}\Url}\fi

\bibitem[Akrour et~al.(2021)Akrour, Tateo, and Peters]{akrour2021continuous}
Riad Akrour, Davide Tateo, and Jan Peters.
\newblock Continuous action reinforcement learning from a mixture of interpretable experts.
\newblock \emph{IEEE Transactions on Pattern Analysis and Machine Intelligence}, 44\penalty0 (10):\penalty0 6795--6806, 2021.

\bibitem[B{\"a}uerle \& Ott(2011)B{\"a}uerle and Ott]{bauerle2011markov}
Nicole B{\"a}uerle and Jonathan Ott.
\newblock Markov decision processes with average-value-at-risk criteria.
\newblock \emph{Mathematical Methods of Operations Research}, 74:\penalty0 361--379, 2011.

\bibitem[Bellemare et~al.(2017)Bellemare, Dabney, and Munos]{bellemare2017distributional}
Marc~G Bellemare, Will Dabney, and R{\'e}mi Munos.
\newblock A distributional perspective on reinforcement learning.
\newblock In \emph{Proceedings of the International Conference on Machine Learning (ICML)}, pp.\  449--458. PMLR, 2017.

\bibitem[Borkar(2002)]{borkar2002q}
Vivek~S Borkar.
\newblock Q-learning for risk-sensitive control.
\newblock \emph{Mathematics of operations research}, 27\penalty0 (2):\penalty0 294--311, 2002.

\bibitem[Brandfonbrener et~al.(2021)Brandfonbrener, Whitney, Ranganath, and Bruna]{brandfonbrener2021offline}
David Brandfonbrener, William~F Whitney, Rajesh Ranganath, and Joan Bruna.
\newblock Offline {RL} without off-policy evaluation.
\newblock In \emph{Advances in Neural Information Processing Systems}, 2021.

\bibitem[Brockman et~al.(2016)Brockman, Cheung, Pettersson, Schneider, Schulman, Tang, and Zaremba]{brockman2016openai}
Greg Brockman, Vicki Cheung, Ludwig Pettersson, Jonas Schneider, John Schulman, Jie Tang, and Wojciech Zaremba.
\newblock Openai gym.
\newblock \emph{arXiv preprint arXiv:1606.01540}, 2016.

\bibitem[Choi et~al.(2019)Choi, Lee, and Oh]{choi2019distributional}
Yunho Choi, Kyungjae Lee, and Songhwai Oh.
\newblock Distributional deep reinforcement learning with a mixture of gaussians.
\newblock In \emph{2019 International Conference on Robotics and Automation (ICRA)}, pp.\  9791--9797. IEEE, 2019.

\bibitem[Chow et~al.(2015)Chow, Tamar, Mannor, and Pavone]{chow2015risk}
Yinlam Chow, Aviv Tamar, Shie Mannor, and Marco Pavone.
\newblock Risk-sensitive and robust decision-making: a cvar optimization approach.
\newblock \emph{Advances in neural information processing systems}, 28, 2015.

\bibitem[Chow et~al.(2018)Chow, Ghavamzadeh, Janson, and Pavone]{chow2018risk}
Yinlam Chow, Mohammad Ghavamzadeh, Lucas Janson, and Marco Pavone.
\newblock Risk-constrained reinforcement learning with percentile risk criteria.
\newblock \emph{Journal of Machine Learning Research}, 18\penalty0 (167):\penalty0 1--51, 2018.

\bibitem[Dabney et~al.(2018{\natexlab{a}})Dabney, Ostrovski, Silver, and Munos]{dabney2018implicit}
Will Dabney, Georg Ostrovski, David Silver, and R{\'e}mi Munos.
\newblock Implicit quantile networks for distributional reinforcement learning.
\newblock In \emph{Proceedings of the International conference on machine learning (ICML)}, pp.\  1096--1105. PMLR, 2018{\natexlab{a}}.

\bibitem[Dabney et~al.(2018{\natexlab{b}})Dabney, Rowland, Bellemare, and Munos]{dabney2018distributional}
Will Dabney, Mark Rowland, Marc Bellemare, and R{\'e}mi Munos.
\newblock Distributional reinforcement learning with quantile regression.
\newblock In \emph{Proceedings of the AAAI Conference on Artificial Intelligence (AAAI)}, volume~32, 2018{\natexlab{b}}.

\bibitem[Daniel et~al.(2012)Daniel, Neumann, and Peters]{daniel2012hierarchical}
Christian Daniel, Gerhard Neumann, and Jan Peters.
\newblock Hierarchical relative entropy policy search.
\newblock In \emph{Artificial Intelligence and Statistics}, pp.\  273--281. PMLR, 2012.

\bibitem[Delbaen \& Biagini(2000)Delbaen and Biagini]{delbaen2000coherent}
Freddy Delbaen and Sara Biagini.
\newblock \emph{Coherent risk measures}.
\newblock Springer, 2000.

\bibitem[Du et~al.(2023)Du, Wang, and Huang]{du2023provably}
Yihan Du, Siwei Wang, and Longbo Huang.
\newblock Provably efficient risk-sensitive reinforcement learning: Iterated cvar and worst path.
\newblock In \emph{Proceedings of the Eleventh International Conference on Learning Representations, {ICLR-23}}, 2023.

\bibitem[Fei et~al.(2021)Fei, Yang, and Wang]{fei2021risk}
Yingjie Fei, Zhuoran Yang, and Zhaoran Wang.
\newblock Risk-sensitive reinforcement learning with function approximation: A debiasing approach.
\newblock In \emph{International Conference on Machine Learning}, pp.\  3198--3207. PMLR, 2021.

\bibitem[Fujimoto \& Gu(2021)Fujimoto and Gu]{fujimoto2021minimalist}
Scott Fujimoto and Shixiang~Shane Gu.
\newblock A minimalist approach to offline reinforcement learning.
\newblock \emph{Advances in neural information processing systems}, 34:\penalty0 20132--20145, 2021.

\bibitem[Fujimoto et~al.(2019)Fujimoto, Meger, and Precup]{fujimoto2019offline}
Scott Fujimoto, David Meger, and Doina Precup.
\newblock Off-policy deep reinforcement learning without exploration.
\newblock \emph{International Conference on Machine Learning}, pp.\  2052--2062, 2019.

\bibitem[Greenberg et~al.(2022)Greenberg, Chow, Ghavamzadeh, and Mannor]{greenberg2022efficient}
Ido Greenberg, Yinlam Chow, Mohammad Ghavamzadeh, and Shie Mannor.
\newblock Efficient risk-averse reinforcement learning.
\newblock \emph{Advances in Neural Information Processing Systems}, 35:\penalty0 32639--32652, 2022.

\bibitem[Haarnoja et~al.(2018)Haarnoja, Zhou, Abbeel, and Levine]{haarnoja2018soft}
Tuomas Haarnoja, Aurick Zhou, Pieter Abbeel, and Sergey Levine.
\newblock Soft actor-critic: Off-policy maximum entropy deep reinforcement learning with a stochastic actor.
\newblock In \emph{International conference on machine learning}, pp.\  1861--1870. PMLR, 2018.

\bibitem[Hau et~al.(2023)Hau, Delage, Ghavamzadeh, and Petrik]{hau2023dynamic}
Jia~Lin Hau, Erick Delage, Mohammad Ghavamzadeh, and Marek Petrik.
\newblock On dynamic programming decompositions of static risk measures in markov decision processes.
\newblock In \emph{Thirty-seventh Conference on Neural Information Processing Systems}, 2023.

\bibitem[Huang et~al.(2021)Huang, Leqi, Lipton, and Azizzadenesheli]{huang2021convergence}
Audrey Huang, Liu Leqi, Zachary~C Lipton, and Kamyar Azizzadenesheli.
\newblock On the convergence and optimality of policy gradient for markov coherent risk.
\newblock \emph{arXiv preprint arXiv:2103.02827}, 2021.

\bibitem[Ji et~al.(2019)Ji, Chang, and Jiang]{ji2019risk}
Ran Ji, KC~Chang, and Zhenlong Jiang.
\newblock Risk-aversion adjusted portfolio optimization with predictive modeling.
\newblock In \emph{2019 22th International Conference on Information Fusion (FUSION)}, pp.\  1--8. IEEE, 2019.

\bibitem[Jung et~al.(2022)Jung, Cho, Park, and Sung]{jung2022quantile}
Whiyoung Jung, Myungsik Cho, Jongeui Park, and Youngchul Sung.
\newblock Quantile constrained reinforcement learning: A reinforcement learning framework constraining outage probability.
\newblock \emph{Advances in Neural Information Processing Systems}, 35:\penalty0 6437--6449, 2022.

\bibitem[Keramati et~al.(2020)Keramati, Dann, Tamkin, and Brunskill]{keramati2020being}
Ramtin Keramati, Christoph Dann, Alex Tamkin, and Emma Brunskill.
\newblock Being optimistic to be conservative: Quickly learning a cvar policy.
\newblock In \emph{Proceedings of the AAAI conference on artificial intelligence}, volume~34, pp.\  4436--4443, 2020.

\bibitem[Kostrikov et~al.(2022)Kostrikov, Nair, and Levine]{kostrikov2022offline}
Ilya Kostrikov, Ashvin Nair, and Sergey Levine.
\newblock Offline reinforcement learning with implicit q-learning.
\newblock In \emph{International Conference on Learning Representations}, 2022.

\bibitem[Kuznetsov et~al.(2020)Kuznetsov, Shvechikov, Grishin, and Vetrov]{kuznetsov2020controlling}
Arsenii Kuznetsov, Pavel Shvechikov, Alexander Grishin, and Dmitry Vetrov.
\newblock Controlling overestimation bias with truncated mixture of continuous distributional quantile critics.
\newblock In \emph{International Conference on Machine Learning}, pp.\  5556--5566. PMLR, 2020.

\bibitem[La \& Ghavamzadeh(2013)La and Ghavamzadeh]{la2013actor}
Prashanth La and Mohammad Ghavamzadeh.
\newblock Actor-critic algorithms for risk-sensitive mdps.
\newblock \emph{Advances in neural information processing systems}, 26, 2013.

\bibitem[Lim \& Malik(2022)Lim and Malik]{lim2022distributional}
Shiau~Hong Lim and Ilyas Malik.
\newblock Distributional reinforcement learning for risk-sensitive policies.
\newblock \emph{Advances in Neural Information Processing Systems}, 35:\penalty0 30977--30989, 2022.

\bibitem[Liu \& Wang(2021)Liu and Wang]{liu2021theory}
Fangda Liu and Ruodu Wang.
\newblock A theory for measures of tail risk.
\newblock \emph{Mathematics of Operations Research}, 46\penalty0 (3):\penalty0 1109--1128, 2021.

\bibitem[Liu et~al.(2022)Liu, Luo, Gaurav, Rezaee, and Poupart]{liu2022benchmarking}
Guiliang Liu, Yudong Luo, Ashish Gaurav, Kasra Rezaee, and Pascal Poupart.
\newblock Benchmarking constraint inference in inverse reinforcement learning.
\newblock \emph{arXiv preprint arXiv:2206.09670}, 2022.

\bibitem[Luo et~al.(2022)Luo, Liu, Duan, Schulte, and Poupart]{luo2022distributional}
Yudong Luo, Guiliang Liu, Haonan Duan, Oliver Schulte, and Pascal Poupart.
\newblock Distributional reinforcement learning with monotonic splines.
\newblock In \emph{International Conference on Learning Representations}, 2022.

\bibitem[Luo et~al.(2023)Luo, Liu, Poupart, and Pan]{luo2024alternative}
Yudong Luo, Guiliang Liu, Pascal Poupart, and Yangchen Pan.
\newblock An alternative to variance: Gini deviation for risk-averse policy gradient.
\newblock \emph{Advances in Neural Information Processing Systems}, 36, 2023.

\bibitem[Malik et~al.(2021)Malik, Anwar, Aghasi, and Ahmed]{malik2021inverse}
Shehryar Malik, Usman Anwar, Alireza Aghasi, and Ali Ahmed.
\newblock Inverse constrained reinforcement learning.
\newblock In \emph{International conference on machine learning}, pp.\  7390--7399. PMLR, 2021.

\bibitem[Mulligan et~al.(2023)Mulligan, Baid, Doctor, Phelps, and Lakdawalla]{mulligan2023risk}
Karen Mulligan, Drishti Baid, Jason~N Doctor, Charles~E Phelps, and Darius~N Lakdawalla.
\newblock Risk preferences over health: Empirical estimates and implications for healthcare decision-making.
\newblock Technical report, National Bureau of Economic Research, 2023.

\bibitem[Osa et~al.(2023)Osa, Hayashi, Deo, Morihira, and Yoshiike]{osa2023offline}
Takayuki Osa, Akinobu Hayashi, Pranav Deo, Naoki Morihira, and Takahide Yoshiike.
\newblock Offline reinforcement learning with mixture of deterministic policies.
\newblock \emph{Transactions on Machine Learning Research}, 2023.

\bibitem[Peng et~al.(2020)Peng, Kumar, Zhang, and Levine]{peng2020advantage}
Xue~Bin Peng, Aviral Kumar, Grace Zhang, and Sergey Levine.
\newblock Advantage weighted regression: Simple and scalable off-policy reinforcement learning.
\newblock \emph{arXiv preprint arXiv:1910.00177}, 2020.

\bibitem[Peters \& Schaal(2007)Peters and Schaal]{peters2007reinforcement}
Jan Peters and Stefan Schaal.
\newblock Reinforcement learning by reward-weighted regression for operational space control.
\newblock In \emph{Proceedings of the 24th international conference on Machine learning}, pp.\  745--750, 2007.

\bibitem[Pflug \& Pichler(2016)Pflug and Pichler]{pflug2016time}
Georg~Ch Pflug and Alois Pichler.
\newblock Time-consistent decisions and temporal decomposition of coherent risk functionals.
\newblock \emph{Mathematics of Operations Research}, 41\penalty0 (2):\penalty0 682--699, 2016.

\bibitem[Puterman(2014)]{puterman2014markov}
Martin~L Puterman.
\newblock \emph{Markov decision processes: discrete stochastic dynamic programming}.
\newblock John Wiley \& Sons, 2014.

\bibitem[Rockafellar et~al.(2000)Rockafellar, Uryasev, et~al.]{rockafellar2000optimization}
R~Tyrrell Rockafellar, Stanislav Uryasev, et~al.
\newblock Optimization of conditional value-at-risk.
\newblock \emph{Journal of risk}, 2:\penalty0 21--42, 2000.

\bibitem[Ruszczy{\'n}ski(2010)]{ruszczynski2010risk}
Andrzej Ruszczy{\'n}ski.
\newblock Risk-averse dynamic programming for markov decision processes.
\newblock \emph{Mathematical programming}, 125:\penalty0 235--261, 2010.

\bibitem[Seyde et~al.(2022)Seyde, Schwarting, Gilitschenski, Wulfmeier, and Rus]{seyde2022strength}
Tim Seyde, Wilko Schwarting, Igor Gilitschenski, Markus Wulfmeier, and Daniela Rus.
\newblock Strength through diversity: Robust behavior learning via mixture policies.
\newblock In \emph{Conference on Robot Learning}, pp.\  1144--1155. PMLR, 2022.

\bibitem[Silver et~al.(2014)Silver, Lever, Heess, Degris, Wierstra, and Riedmiller]{silver2014deterministic}
David Silver, Guy Lever, Nicolas Heess, Thomas Degris, Daan Wierstra, and Martin Riedmiller.
\newblock Deterministic policy gradient algorithms.
\newblock In \emph{International conference on machine learning}, pp.\  387--395. Pmlr, 2014.

\bibitem[Sutton \& Barto(1998)Sutton and Barto]{sutton1998introduction}
Richard~S. Sutton and Andrew~G. Barto.
\newblock \emph{Reinforcement Learning: {A}n Introduction}.
\newblock The MIT Press, Cambridge, MA, 1998.

\bibitem[Tamar et~al.(2012)Tamar, Di~Castro, and Mannor]{tamar2012policy}
Aviv Tamar, Dotan Di~Castro, and Shie Mannor.
\newblock Policy gradients with variance related risk criteria.
\newblock In \emph{Proceedings of the twenty-ninth international conference on machine learning}, pp.\  387--396, 2012.

\bibitem[Tamar et~al.(2015)Tamar, Glassner, and Mannor]{tamar2015optimizing}
Aviv Tamar, Yonatan Glassner, and Shie Mannor.
\newblock Optimizing the cvar via sampling.
\newblock In \emph{Proceedings of the AAAI Conference on Artificial Intelligence}, volume~29, 2015.

\bibitem[Tang et~al.(2019)Tang, Zhang, and Salakhutdinov]{tang2019worst}
Yichuan~Charlie Tang, Jian Zhang, and Ruslan Salakhutdinov.
\newblock Worst cases policy gradients.
\newblock \emph{Conference on Robot Learning}, 2019.

\bibitem[Todorov et~al.(2012)Todorov, Erez, and Tassa]{todorov2012mujoco}
Emanuel Todorov, Tom Erez, and Yuval Tassa.
\newblock Mujoco: A physics engine for model-based control.
\newblock In \emph{2012 IEEE/RSJ international conference on intelligent robots and systems}, pp.\  5026--5033. IEEE, 2012.

\bibitem[Wulfmeier et~al.(2020)Wulfmeier, Abdolmaleki, Hafner, Springenberg, Neunert, Hertweck, Lampe, Siegel, Heess, and Riedmiller]{wulfmeier2020compositional}
Markus Wulfmeier, Abbas Abdolmaleki, Roland Hafner, Jost~Tobias Springenberg, Michael Neunert, Tim Hertweck, Thomas Lampe, Noah Siegel, Nicolas Heess, and Martin Riedmiller.
\newblock Compositional transfer in hierarchical reinforcement learning.
\newblock \emph{Robotics: Science and Systems}, 2020.

\bibitem[Wulfmeier et~al.(2021)Wulfmeier, Rao, Hafner, Lampe, Abdolmaleki, Hertweck, Neunert, Tirumala, Siegel, Heess, et~al.]{wulfmeier2021data}
Markus Wulfmeier, Dushyant Rao, Roland Hafner, Thomas Lampe, Abbas Abdolmaleki, Tim Hertweck, Michael Neunert, Dhruva Tirumala, Noah Siegel, Nicolas Heess, et~al.
\newblock Data-efficient hindsight off-policy option learning.
\newblock In \emph{International Conference on Machine Learning}, pp.\  11340--11350. PMLR, 2021.

\bibitem[Xiao et~al.(2023)Xiao, Wang, Pan, White, and White]{xiao2023offline}
Chenjun Xiao, Han Wang, Yangchen Pan, Adam White, and Martha White.
\newblock The in-sample softmax for offline reinforcement learning.
\newblock \emph{International Conference on Learning Representations}, 2023.

\bibitem[Yang et~al.(2021)Yang, Sim{\~a}o, Tindemans, and Spaan]{yang2021wcsac}
Qisong Yang, Thiago~D Sim{\~a}o, Simon~H Tindemans, and Matthijs~TJ Spaan.
\newblock Wcsac: Worst-case soft actor critic for safety-constrained reinforcement learning.
\newblock In \emph{Proceedings of the AAAI Conference on Artificial Intelligence}, volume~35, pp.\  10639--10646, 2021.

\bibitem[Ying et~al.(2022)Ying, Zhou, Su, Yan, Chen, and Zhu]{ying2022towards}
ChengYang Ying, Xinning Zhou, Hang Su, Dong Yan, Ning Chen, and Jun Zhu.
\newblock Towards safe reinforcement learning via constraining conditional value-at-risk.
\newblock In \emph{Proceedings of the Thirty-First International Joint Conference on Artificial Intelligence, {IJCAI-22}}, pp.\  3673--3680, 7 2022.

\bibitem[Yu et~al.(2023)Yu, Wang, and Dong]{yu2023learning}
Shi Yu, Haoran Wang, and Chaosheng Dong.
\newblock Learning risk preferences from investment portfolios using inverse optimization.
\newblock \emph{Research in International Business and Finance}, 64:\penalty0 101879, 2023.

\bibitem[Yue et~al.(2020)Yue, Wang, and Zhou]{yue2020implicit}
Yuguang Yue, Zhendong Wang, and Mingyuan Zhou.
\newblock Implicit distributional reinforcement learning.
\newblock \emph{Advances in Neural Information Processing Systems}, 33:\penalty0 7135--7147, 2020.

\bibitem[Zhang \& Whiteson(2019)Zhang and Whiteson]{zhang2019dac}
Shangtong Zhang and Shimon Whiteson.
\newblock Dac: The double actor-critic architecture for learning options.
\newblock \emph{Advances in Neural Information Processing Systems}, 32, 2019.

\bibitem[Zhou et~al.(2020)Zhou, Wang, and Feng]{NEURIPS2020_b6f8dc08}
Fan Zhou, Jianing Wang, and Xingdong Feng.
\newblock Non-crossing quantile regression for distributional reinforcement learning.
\newblock In H.~Larochelle, M.~Ranzato, R.~Hadsell, M.~F. Balcan, and H.~Lin (eds.), \emph{Advances in Neural Information Processing Systems (NeurIPS)}, volume~33, pp.\  15909--15919. Curran Associates, Inc., 2020.

\bibitem[Zhou et~al.(2021)Zhou, Zhu, Kuang, and Zhang]{ijcai2021-476}
Fan Zhou, Zhoufan Zhu, Qi~Kuang, and Liwen Zhang.
\newblock Non-decreasing quantile function network with efficient exploration for distributional reinforcement learning.
\newblock In \emph{Proceedings of International Joint Conference on Artificial Intelligence (IJCAI)}, pp.\  3455--3461. International Joint Conferences on Artificial Intelligence Organization, 2021.

\end{thebibliography}
\bibliographystyle{rlc}

\newpage
\appendix

\section{Experiments Details}
\label{sec:experiments}
\subsection{General Descriptions of Different Methods}
\label{sec:baseline-info}
\textbf{Policy gradient methods.} Among the methods included in this paper, REINFOCE~\citep{sutton1998introduction}, CVaR-PG~\citep{tamar2015optimizing}, and the CVaR part of our mixture policy are on-policy policy gradient methods. In our implementation, we update the policy after gathering $N$ trajectories.

\textbf{Time difference methods.} DRL-mkv~\citep{dabney2018implicit,keramati2020being}, DRL-lim~\citep{lim2022distributional}, SAC~\citep{haarnoja2018soft} are off-policy time difference methods, i.e., updating policy at each environment step. Continuous action domains are not considered in the original paper of~\citet{lim2022distributional}. We follow~\citet{tang2019worst} to apply DRL-mkv and DRL-lim to continuous action domains, e.g., for DRL-mkv, the actor is updated via
\begin{equation}
    \nabla_\theta J_\alpha = \mathbb{E}_{s,a\sim\pi_\theta} [\nabla_\theta \log\pi_\theta(a|s)\mathrm{CVaR}_\alpha(Z^\pi(s,a))]
\end{equation}
for DRL-lim, the actor is updated via
\begin{equation}
    \nabla_\theta J_\alpha = \mathbb{E}_{s,k,a\sim\pi_\theta}\Big[\nabla_\theta -\log\pi(a|s)\mathbb{E}[(k-Z^\pi(s,a))^+]\Big]
\end{equation}
where $k$ is the tracking variable at state $s$.

Both DRL-mkv and DRL-lim are built on top of distributional RL~\citep{bellemare2017distributional}. The most commonly used approach to update distributional value function (critic) is quantile regression~\citep{dabney2018distributional,dabney2018implicit,NEURIPS2020_b6f8dc08,ijcai2021-476,luo2022distributional}. We also adopt quantile regression in our implementation. 

\textbf{Solving the quantile crossing issue.} In particular, \citet{NEURIPS2020_b6f8dc08} pointed out some previous quantile regression based work, e.g., QR-DQN~\citep{dabney2018distributional}, IQN~\citep{dabney2018implicit} suffered from the quantile crossing issue, i.e., the predicted quantile values do not satisfy the monotonicity of the quantile function. This is shown to hinder policy learning and exploration~\citep{NEURIPS2020_b6f8dc08}. The monotonicity of the quantile is also important in DRL-mkv and DRL-lim to make sure the estimated quantities, e.g. $\alpha$-CVaR, are correct. We follow the approach in~\citet{yue2020implicit} by sorting the predicted quantile values to make them non decreasing.

\subsection{The Maze Problem}
The maze consists of a $6\times 6$ grid. The initial state of the agent is fixed at the bottom left corner. The action space is four (up, down, left, right). The maximum episode length is $100$.

\textbf{Policy function.} For CVaR-PG, the policy is represented as
\begin{equation}
   \pi_\theta(a|s) = \frac{e^{\zeta(s,a)\cdot\theta}}{\sum_b e^{\zeta(s,b)\cdot\theta}}
\end{equation}
where $\zeta(s,a)$ is the state-action feature vector, basically a one-hot encoding in our implementation. Thus, the dimension of $\zeta(s,a)$ is $6\times 6 \times 4$. The derivative of the logarithm is
\begin{equation}
    \nabla_\theta \log \pi_\theta(a|s) = \zeta(s,a)-\mathbb{E}_{b\sim \pi_\theta(\cdot|s)}\zeta(s,b)
\end{equation}

For our mixture policy, the policy parameter $\theta$ consists of two parts $\theta=(\theta_1, \theta_2)$, where $\theta_1$ is for the adjustive policy $\pi'_{\theta_1}$, and $\theta_2$ is for the weight $w$.
\begin{align}
    \pi_\theta(a|s)=\sigma(\zeta(s,a)\cdot\theta_2)&\pi'_{\theta_1}(a|s) + \Big(1-\sigma(\zeta(s,a)\cdot\theta_2)\Big)\pi^n(a|s)\\
    \mathrm{with~} &\pi'_{\theta_1}=\frac{e^{\zeta(s,a)\cdot\theta_1}}{\sum_b e^{\zeta(s,b)\cdot\theta_1}}
\end{align}
where $\sigma(\cdot)$ is the sigmoid function. The derivative of the logarithm is
\begin{align}
    \nabla_{\theta_1}\log \pi_\theta(a|s)&= \frac{1}{\pi_\theta(a|s)}\sigma(\zeta(s,a)\cdot\theta_2)\pi'_{\theta_1}(a|s)\nabla_{\theta_1}\log\pi'_{\theta_1}(a|s)\\
    \nabla_{\theta_2}\log \pi_\theta(a|s)&= \frac{1}{\pi_\theta(a|s)}(\pi'_{\theta_1}(a|s)-\pi^n(a|s))\sigma(\zeta(s,a)\cdot\theta_2)\Big(1-\sigma(\zeta(s,a)\cdot\theta_2)\Big)\zeta(s,a)
\end{align}

\textbf{Value function.} The value function of REINFORCE baseline is represented as $V_\upsilon(s)=\zeta(s)\cdot \upsilon$. Similarly, $\zeta(s)$ is a one-hot encoding.

\subsubsection{Learning Parameters}
Discount factor $\gamma=0.999$. Optimizer is stochastic gradient descent (SGD).

\textbf{REINFORCE}: Policy learning rate is  1e-2$\in\{$1e-2, 1e-3, 1e-4$\}$. Value leaning rate is 10 times policy learning rate. (the suffix 'e-2' means 0.01)

\textbf{CVaR-PG}: Learning rate is  1e-2$\in\{$1e-1, 1e-2, 1e-3, 1e-4$\}$.

\textbf{MIX}: Learning rate is  1e-2$\in\{$1e-2, 1e-3, 1e-4$\}$.

\subsubsection{Policy Gradient Norm of CVaR-PG}

\begin{figure}[ht]
    \begin{center}
        \includegraphics[width=0.35\textwidth]{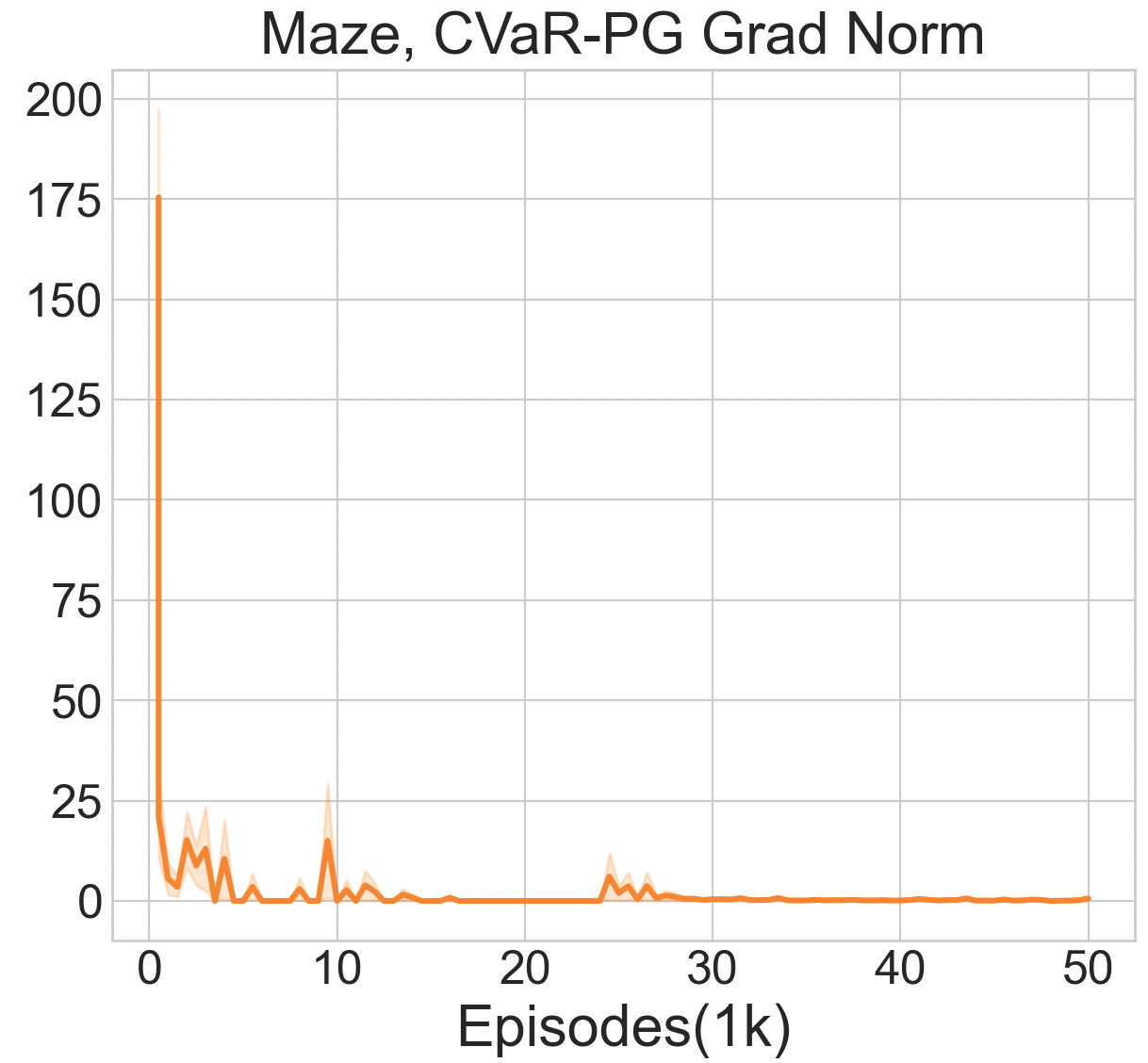}
    \end{center}
    \caption{Policy gradient norm (y-axsis) of CVaR-PG in Maze. Curves are averaged over 10 seeds with shaded regions indicating standard errors}
    \label{fig:cvarpg-norm}
\end{figure}

We show the policy gradient norm of CVaR-PG to further demonstrate the gradient vanishing issue in Maze. The norm is computed by \texttt{numpy.linalg.norm}. As shown in Fig.~\ref{fig:cvarpg-norm}, in most of the time, the policy gradients are zero.

\subsection{LunarLander Discrete}
\begin{figure}[ht]
    \begin{center}
        \includegraphics[width=0.3\textwidth]{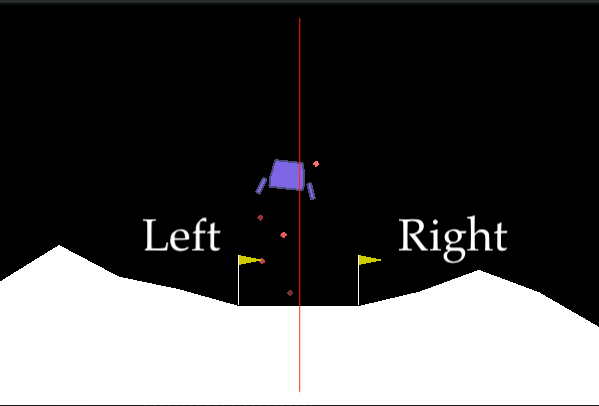}
    \end{center}
    \caption{Split the ground of LunarLander into left and right parts by the middle (red) line. If landing on the right, an additional reward smapled from $\mathcal{N}(0,1)$ times $100$ is given.}
    \label{fig:LL}
\end{figure}

The goal is to land the agent on the ground without crashing. The state dimension is $8$. The action dimension is $4$. The detailed reward information is available at this webpage~\footnote{https://www.gymlibrary.dev/environments/box2d/lunar\_lander/}. Here, we split the ground into left and right parts by the middle line of the landing pad as shown in Figure~\ref{fig:LL}. If the agent lands on the right part of the ground, an additional noisy reward signal $\mathcal{N}(0, 1)\times 100$ is given. The maximum episode length is $500$.

\textbf{Policy function.} The policy is a categorical distribution in REINFORCE and CVaR-PG, modeled as a neutral network. Hidden layer: 2. Hidden size: 128. Activation: ReLU. Softmax function is applied to the output to generate categorical probabilities.

The policy of MIX is a weighted summation of $\pi'$ and $\pi^n$ with weight $w$. $\pi'$ and $w$ are modeled as a neutral network with two output heads. $\pi^n$ is a separate neutral network. Both of them have: Hidden layer: 2. Hidden size: 128. Activation: ReLU.

\textbf{Value function.} For value function in REINFORCE baseline, $Q$ and $V$ function in IQL of MIX. Hidden layer: 2. Hidden size: 128. Activation: ReLU.

For distributional value function in DRL-mkv and DRL-lim. Hidden layer: 2. Hidden size: 128. Activation: ReLU. Quantile size (i.e., final layer output size): 80.

\subsubsection{Learning Parameters}

Discount factor $\gamma=0.999$. Optimizer is Adam.

\textbf{REINFORCE}: Policy learning rate is  7e-4$\in\{$1e-3, 7e-4, 3e-4, 1e-4$\}$. Value leaning rate is 10 times policy learning rate.

\textbf{CVaR-PG}: Learning rate is  7e-4$\in\{$1e-3, 7e-4, 3e-4, 1e-4$\}$.

\textbf{MIX}: Learning rate for $\pi'$ and $w$ is 7e-4$\in\{$1e-3, 7e-4, 3e-4, 1e-4$\}$. Learning rate for IQL part (including policy and value functions) is 1e-4$\in\{$3e-4, 1e-4$\}$. IQL update frequency $C=50$, by sampling 2e5 transitions from buffer. $\eta=0.8$ in Eq.~\ref{eq:iql-v}. $\beta=1$ in Eq.~\ref{eq:iql-pi}.

\textbf{DRL-mkv}: Learning rate is 7e-4$\in\{$1e-3, 7e-4, 3e-4, 1e-4, 7e-5$\}$.

\textbf{DRL-lim}: Learning rate is 1e-4$\in\{$1e-3, 7e-4, 3e-4, 1e-4, 7e-5$\}$.

\subsubsection{Performance of the risk neutral component of the mixture policy}

\begin{figure}[ht]
    \begin{center}
        \includegraphics[width=0.3\textwidth]{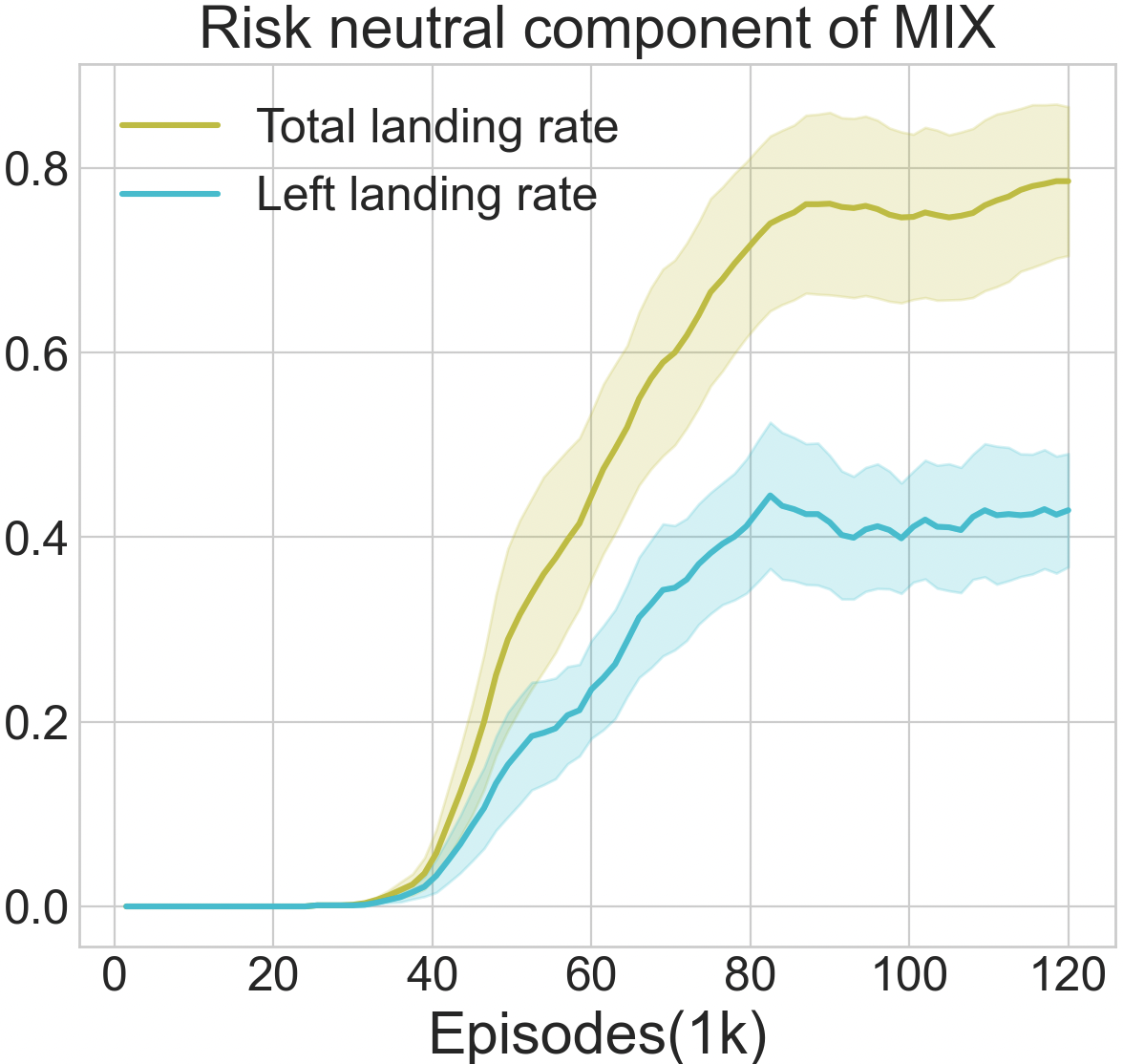}
    \end{center}
    \caption{Total landing rate and left landing rate of IQL (y-axsis) in MIX during training. Curves are averaged over 10 seeds with shaded regions indicating
standard errors.}
    \label{fig:ll-iql}
\end{figure}

In this domain, the risk neutral component $\pi^n$ of the mixture policy is updated vi IQL~\citep{kostrikov2022offline}. We report the total successful landing rate and left landing rate of $\pi^n$ during training in Fig.~\ref{fig:ll-iql}. $\pi^n$ does not demonstrates a preference of landing location, which indicates the risk-aversion is achieved by the mixture policy.

\subsubsection{Incorporate pre-trained risk-neutral policy}
\label{sec:pre-train-risk-neutral}

\begin{figure}[ht]
    \begin{center}
        \includegraphics[width=0.65\textwidth]{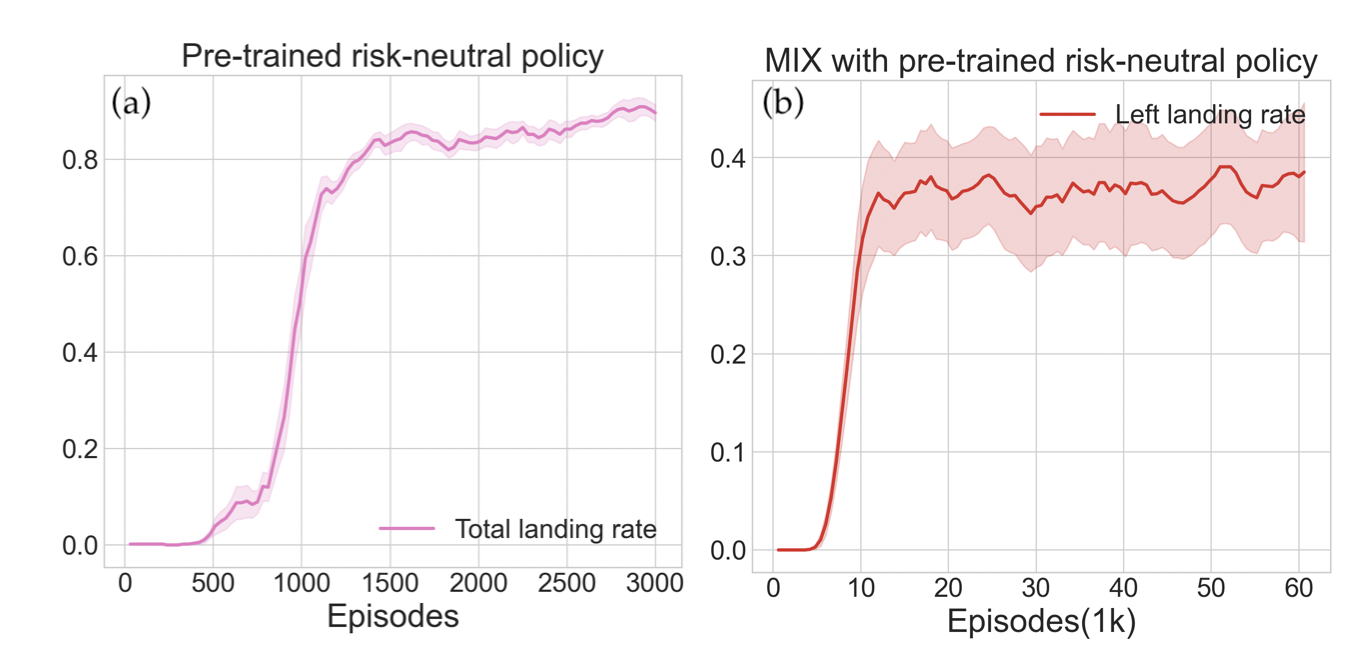}
    \end{center}
    \caption{(a) The total successful landing rate (y-axsis) of pre-trained risk-neutral policy. (b) The left landing (i.e., risk-averse) (y-axsis) rate of Mix by incorporating this pre-trained risk-neutral policy. Curves are averaged over 10 seeds with shaded regions indicating
standard errors.}
    \label{fig:ll-dqn}
\end{figure}

A natural question raises that whether we can use a pre-trained risk-neural policy to form the mixture policy, as done in maze (Sec.~\ref{sec:maze}), when using deep RL. We conduct the experiments in this LunarLander domain.

Similar as the risk-neutral policy in maze, we represent the risk-neutral policy by the softmax of $Q$-values with temperature. The $Q$-values are learned by deep $Q$-network (DQN). To validate the pre-trained risk-neutral policy performs well, we show its total successful landing rate in Fig.~\ref{fig:ll-dqn}(a). The whole training process for MIX is as follows. The first 3k episodes are used to update the risk-neutral policy only (i.e., update DQN), with the remaining part of the MIX policy unchanged. After the first 3k episodes, the risk-neutral policy is fixed, and the remaning part of the MIX policy begins to update. However, as indicated by the left landing rate in Fig.~\ref{fig:ll-dqn}(b), MIX leads to a suboptimal risk-averse policy. One possible reason may be when training the deep risk-neutral RL algorithm, the data distribution tends to concentrate on those states in the optimal (or near optimal) trajectories. Thus, the learned value or policy function approximator may not generalize well around the risk-averse path. 

\subsection{InvertedPendulum}
The description of the Mujoco environments can be found at this webpage~\footnote{https://www.gymlibrary.dev/environments/mujoco/}.

The goal is to balance a inverted pendulum on a cart. The state dimension is 4 (X-position is already contained in the observation). The action dimension is 1. Per step reward is 1. If the agent reaches the region X-position > 0.04, and additional noisy reward sampled from $\mathcal{N}(0,1)$ times 10 is given. The game ends if angle between the pendulum and the cart is greater than 0.2 radian or a maximum episode length 300 is reached.

\textbf{Policy function.} The policy is a normal distribution in REINFORCE and CVaR-PG, modeled as a neutral network. Hidden layer: 2. Hidden size: 128. Activation: ReLU. Tanh is applied to the last layer. The logarithm of standard deviation is an independent trainable parameter.

For MIX, $\pi'$ and $w$ is a neutral network with two output heads. One for the mean of the normal distribution $\pi'$, one for $w$. The logarithm of standard deviation is an independent trainable parameter. $\pi^n$ is a seperate neutral network as above. Both of them have: Hidden layer: 2. Hidden size: 128. Activation: ReLU. Tanh is applied to the output of the distribution layer.

\textbf{Value function.} For value function in REINFORCE baseline, $Q$ and $V$ function in IQL of MIX. Hidden layer: 2. Hidden size: 128. Activation: ReLU.

For distributional value function in DRL-mkv and DRL-lim. Hidden layer: 2. Hidden size: 128. Activation: ReLU. Quantile size (i.e., final layer output size): 80.

\subsubsection{Learning Parameters}

Discount factor $\gamma=0.999$. Optimizer is Adam.

\textbf{REINFORCE}: Policy learning rate is  3e-4$\in\{$7e-4, 3e-4, 1e-4$\}$. Value leaning rate is 10 times policy learning rate.

\textbf{CVaR-PG}: Learning rate is  3e-4$\in\{$7e-4, 3e-4, 1e-4$\}$.

\textbf{MIX}: Learning rate for $\pi'$ and $w$ is 3e-4$\in\{$7e-4, 3e-4, 1e-4$\}$. Learning rate for IQL part (including policy and value functions) is 1e-4$\in\{$3e-4, 1e-4$\}$. IQL update frequency $C=50$, by sampling 1e5 transitions from buffer. $\eta=0.9$ in Eq.~\ref{eq:iql-v}. $\beta=2$ in Eq.~\ref{eq:iql-pi}.

\textbf{DRL-mkv}: Learning rate is 7e-4$\in\{$1e-3, 7e-4, 3e-4, 1e-4, 7e-5$\}$.

\textbf{DRL-lim}: Learning rate is 1e-3$\in\{$1e-3, 7e-4, 3e-4, 1e-4, 7e-5$\}$.

\subsection{HalfCheetah}
The agent controls a robot with two legs. The state dimension is 18 (add X-position). The action dimension is 6. One part of the reward is determined by the distance covered between the current and the previous time step. Originally, it is positive only when the agent moves toward the forward (right) direction. To encourage the agent to freely move forward (left) and backward (right), we modify this part of the reward to make it positive as long as the agent is moving far from the origin. If the agent reaches the region X-position $<$-3, an additional noisy reward sampled from $\mathcal{N}(0,1)$ times 50 is given. The game ends when a maximum episode length 500 is reached.

\textbf{Policy function.} Hidden size: 256. Others are the same as the case in InvertedPendulum.

\textbf{Value function.} Hidden size: 256. Others are the same as the case in InvertedPendulum.

\subsubsection{Learning Parameters}

Discount factor $\gamma=0.99$. Optimizer is Adam.

\textbf{SAC}: Learning rate is  3e-4$\in\{$7e-4, 3e-4, 1e-4$\}$.

\textbf{CVaR-PG}: Learning rate is  3e-4$\in\{$7e-4, 3e-4, 1e-4$\}$.

\textbf{MIX}: Learning rate for $\pi'$ and $w$ is 3e-4$\in\{$7e-4, 3e-4, 1e-4$\}$. Learning rate for IQL part (including policy and value functions) is the same. IQL update frequency $C=30$, by sampling 2e5 transitions from buffer. $\eta=0.8$ in Eq.~\ref{eq:iql-v}. $\beta=2$ in Eq.~\ref{eq:iql-pi}.

\textbf{DRL-mkv}: Learning rate is 3e-4$\in\{$7e-4, 3e-4, 1e-4, 7e-5$\}$.

\textbf{DRL-lim}: Learning rate is 1e-4$\in\{$7e-4, 3e-4, 1e-4, 7e-5$\}$.

\subsubsection{Policy Return in HalfCheetah}

\begin{figure}[ht]
    \begin{center}
        \includegraphics[width=0.6\textwidth]{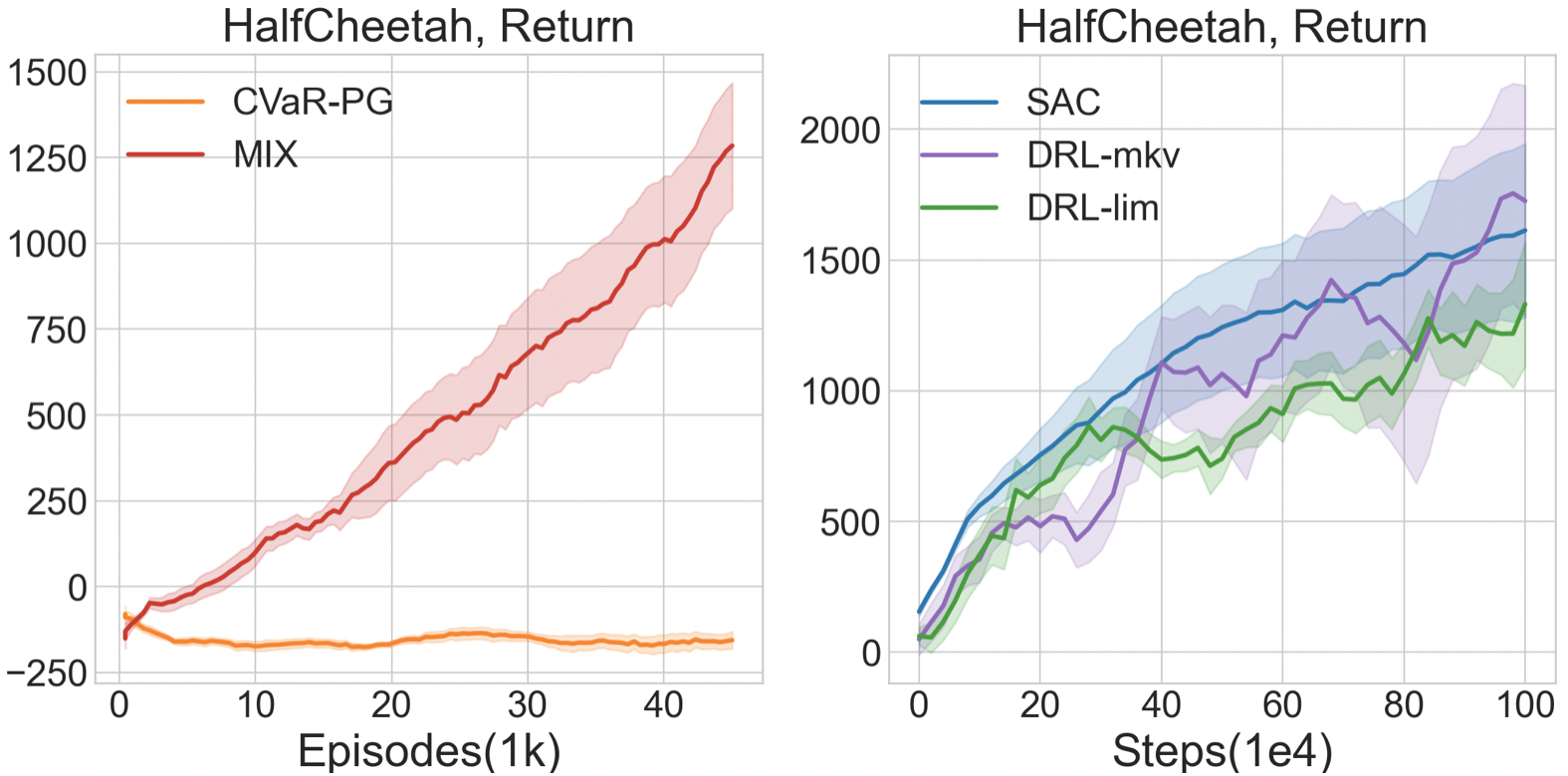}
    \end{center}
    \caption{Policy return (y-axsis) v.s. training episodes or steps in HalfCheetah. Curves are averaged over 10 seeds with shaded regions indicating standard errors.}
    \label{fig:hc-ret}
\end{figure}

The policy return of different methods in HalfCheetah is shown in Fig.~\ref{fig:hc-ret}. Note that the policy return does not indicate the risk-aversion of a policy. 

\subsection{Ant}
The agent controls a robot with four legs attached to it with each leg having two links. The state dimension is 113 (add X-position). The action dimension is 8. Similar to HalfCheetah, we modify the reward to make the distance based reward positive as long as the agent is moving far from the origin. If the agent reaches the region X-position $<$-3, an additional noisy reward sampled from $\mathcal{N}(0,1)$ times 50 is given. The game ends when a maximum episode length 500 is reached.

\textbf{Policy function.} Hidden size: 256. Others are the same as the case in InvertedPendulum.

\textbf{Value function.} Hidden size: 256. Others are the same as the case in InvertedPendulum.

\subsubsection{Learning Parameters}

Discount factor $\gamma=0.99$. Optimizer is Adam.

\textbf{SAC}: Learning rate is  3e-4$\in\{$7e-4, 3e-4, 1e-4$\}$.

\textbf{CVaR-PG}: Learning rate is  3e-4$\in\{$7e-4, 3e-4, 1e-4$\}$.

\textbf{MIX}: Learning rate for $\pi'$ and $w$ is 3e-4$\in\{$7e-4, 3e-4, 1e-4$\}$. Learning rate for IQL part (including policy and value functions) is the same. IQL update frequency $C=30$, by sampling 2e5 transitions from buffer. $\eta=0.8$ in Eq.~\ref{eq:iql-v}. $\beta=2$ in Eq.~\ref{eq:iql-pi}.

\textbf{DRL-mkv}: Learning rate is 7e-5$\in\{$7e-4, 3e-4, 1e-4, 7e-5$\}$.

\textbf{DRL-lim}: Learning rate is 7e-5$\in\{$7e-4, 3e-4, 1e-4, 7e-5$\}$.

\subsubsection{Policy Return in Ant}
\begin{figure}[ht]
    \begin{center}
        \includegraphics[width=0.6\textwidth]{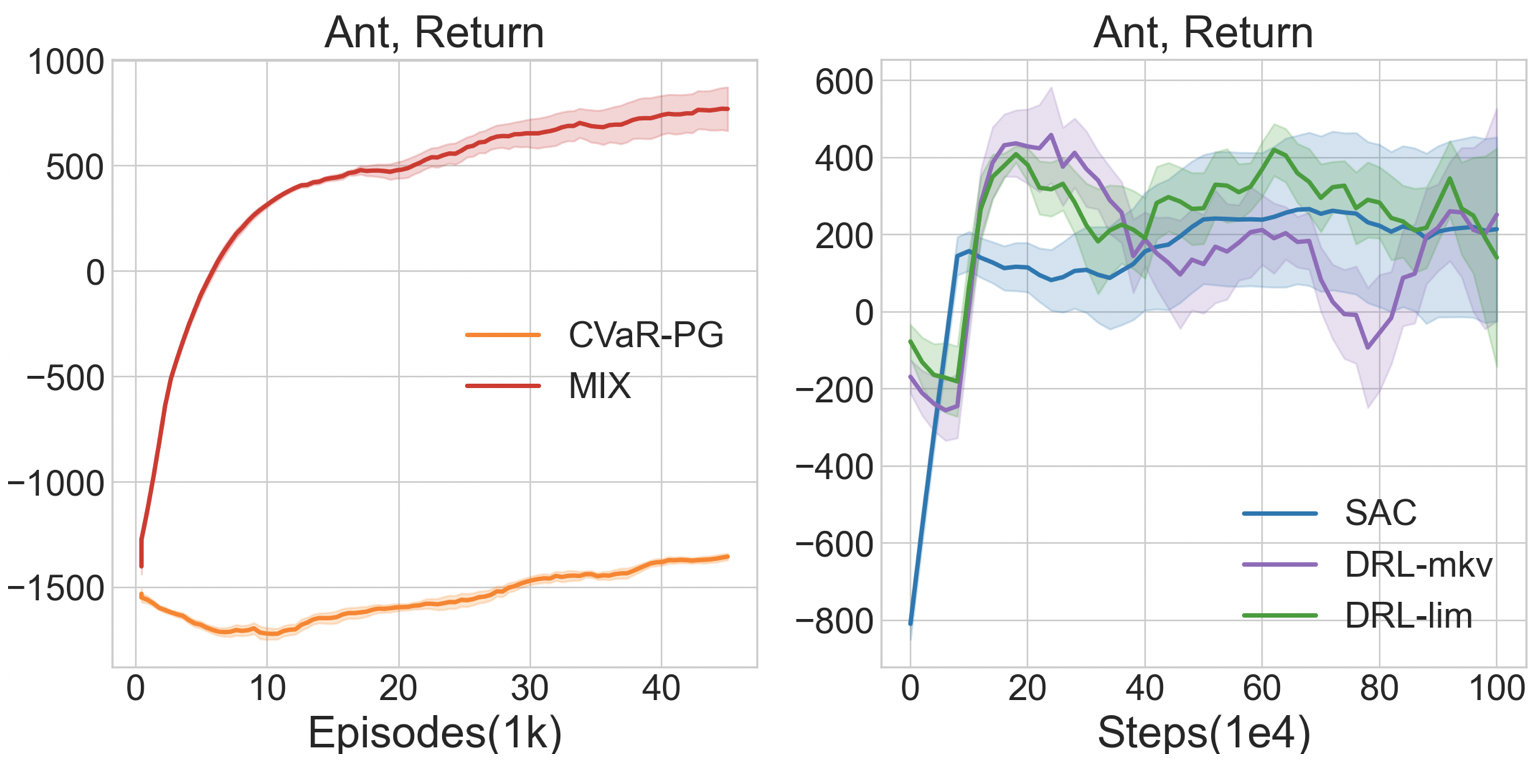}
    \end{center}
    \caption{Policy return (y-axsis) v.s. training episodes or steps in Ant. Curves are averaged over 10 seeds with shaded regions indicating standard errors.}
    \label{fig:ant-ret}
\end{figure}

The policy return of different methods in Ant is shown in Fig.~\ref{fig:ant-ret}. Note that the policy return does not indicate the risk-aversion of a policy. 

\subsection{Driving Game}
\label{sec:driving}
The goal of this game is to control the agent’s car to follow the leader car  without colliding. The state dimension is 5. The action dimension is 5. We refer reader to Sec. 5.2 of~\citep{greenberg2022efficient} for more details.

The method proposed in ~\citet{greenberg2022efficient} is named CeSoR, which includes a curriculum learning scheduler to adjust CVaR $\alpha$ during learning, i.e., starting from a large value for $\alpha$ and gradually decreasing to its target value; and a trajectory generator which controls the environment dynamic. In this domain, it controls the behavior of the leader car.

We directly use the code provided by~\citet{greenberg2022efficient} to produce the results for CeSoR. CeSoR is orthogonal to MIX, and these two can be combined. We combine MIX with curriculum learning (denoted as MIX+SoR, SoR means soft risk to represent curriculum learning in \citet{greenberg2022efficient}), and combine MIX with CeSoR (denoted as MIX+CeSoR) in this domain.

\textbf{Policy function.} Policy of CeSoR and CVaR-PG: Hidden size: 32. Hidden layer: 2. Activation: Tanh.

Policy of MIX: Hidden size: 32. Activation: Tanh. Others are the same as the case in InvertedPendulum.

\textbf{Value function.} $Q$ and $V$ of IQL: Hidden size: 32. Others are the same as the case in InvertedPendulum.

\subsubsection{Learning Parameters}
CVaR $\alpha=0.05$. Update policy after gathering $N=80$ trajectories. The starting value for CVaR $\alpha$ is 0.8.

\textbf{CVaR-PG}: Learning rate 1e-2$\in\{$2e-2, 1e-2, 5e-3$\}$.

\textbf{CeSoR}: Learning rate 1e-2$\in\{$2e-2, 1e-2, 5e-3$\}$.

\textbf{MIX}: Learning rate for $\pi'$ and $w$ is 1e-2$\in\{$2e-2, 1e-2, 5e-3$\}$. Learning rate for IQL part (including policy and value function) is 5e-3. IQL update frequency $C=50$, by sample 2e4 transitions from buffer. $\eta=0.8$ in Eq.~\ref{eq:iql-v}. $\beta=2$ in Eq.~\ref{eq:iql-pi}.

\textbf{MIX+SoR}: Learning parameters are the same as MIX.

\textbf{MIX+CeSoR}: Learning parameters are the same as MIX.

\begin{figure}[ht]
    \begin{center}
        \includegraphics[width=0.73\textwidth]{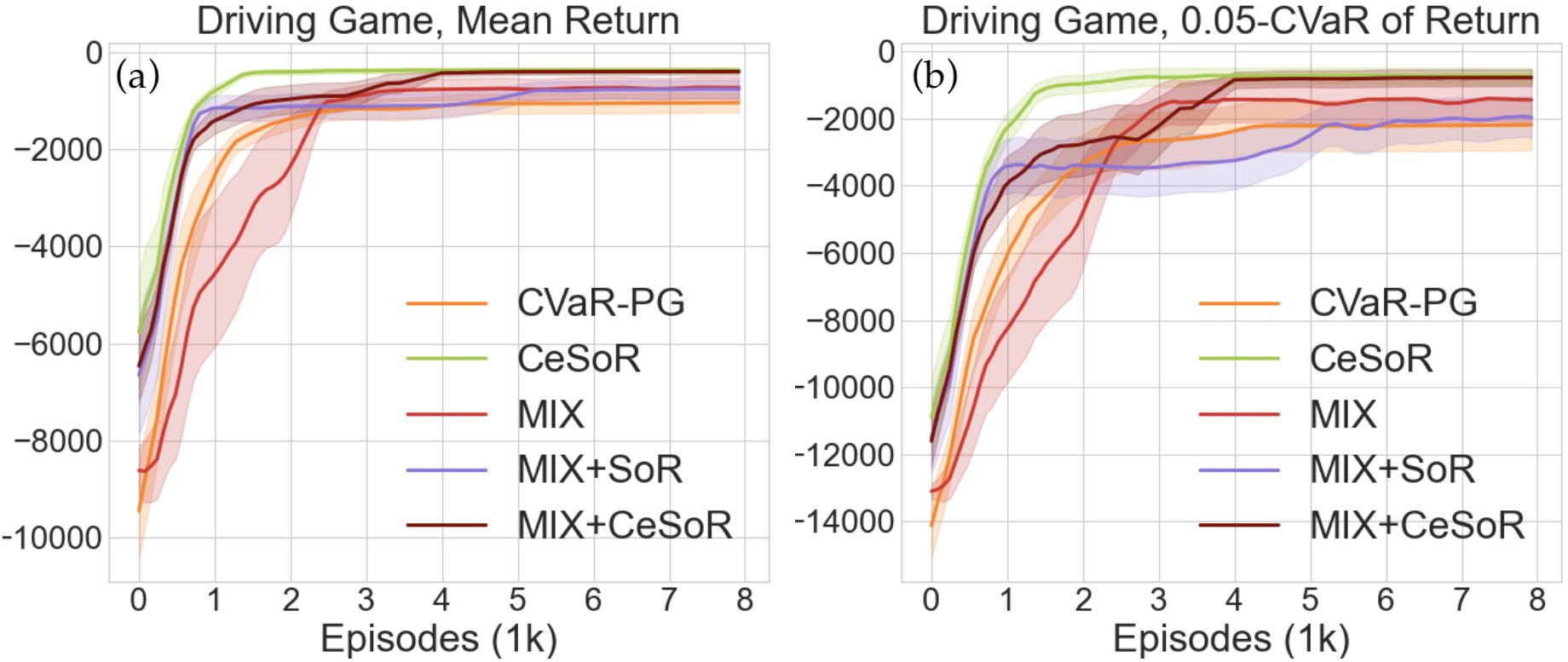}
    \end{center}
    \caption{(a) The expected return (y-axsis), and (b) the 0.05-CVaR of the return (y-axsis) achieved by CVaR-PG, CeSoR, MIX, MIX+SoR, and MIX+CeSoR in driving game. Curves are averaged over 10 seeds with shaded regions indicating standard errors.}
    \label{fig:driving}
\end{figure}

We report the mean return and the 0.05-CVaR of the return in Fig.~\ref{fig:driving}. For both the mean and tail of the return, CeSoR, MIX, and MIX variants are better than CVaR-PG. CeSoR is slightly better than MIX, since CeSoR possesses the environment dynamic information while MIX does not. MIX with 
curriculum learning, i.e., MIX+SoR, learns faster than MIX at the early training stage than MIX, though the final mean return is the same as MIX. MIX+CeSoR is better than MIX and MIX+SoR with respect to the mean and tail of the return, and is comparable to CeSoR. CeSoR learns faster and achieves the highest risk averse rewards among all techniques, however it requires access to the environment dynamics and the ability to change the parameters of the dynamics in a way that is domain specific. In contrast, MIX and MIX+SoR do not need dynamics information and therefore can be applied directly to any domain.  

\textbf{Remark.} Our mixture policy method differs from the curriculum learning idea in~\citet{greenberg2022efficient}. Tough the CVaR $\alpha$ starts from a large value in curriculum learning, where the objective is close to a risk-neutral problem, it is an on-policy policy gradient method, i.e., the trajectories used to update the policy is generated by the current policy (if no importance sampling is assumed). In contrast, the risk-neutral component of our mixture policy is trained by an off-policy (offline) algorithm, in this case, all the encountered trajectories can be stored in the replay buffer for policy update.

\section{Additional Related Work}
    Due to its two-layered structure, a mixture policy is also called a hierarchical policy~\citep{daniel2012hierarchical}. Though the idea of mixture policy is not new, it is mainly applied in risk-neutral settings. \citet{osa2023offline} constructed a mixture of deterministic policies for offline RL tasks and shown it can mitigate the issue of critic error accumulation in offline RL. \cite{wulfmeier2020compositional} and \citet{seyde2022strength} utilized mixture policy to capture the diverse motivations of the robots such that the skill learned by each sub-policy can be transferred. \citet{akrour2021continuous} adopted mixture policy to enhance the interpretability of decision making. The mixture policy is also used for option discovery~\citep{zhang2019dac, wulfmeier2021data}.
The similar mixture structure also appears in value (critic) function learning, for instance, mixture critic is utilized for distributional RL~\citep{choi2019distributional,kuznetsov2020controlling}.

\end{document}